\documentclass[10pt,leqno]{amsart}

\usepackage[utf8]{inputenc}
\usepackage[T1]{fontenc}

\usepackage{geometry}
\AtBeginEnvironment{table}{}
\AfterEndEnvironment{table}{}

\usepackage{amssymb,amsthm,amsmath,amsfonts,nicefrac}

\usepackage{float,graphicx,booktabs,multirow,array,colortbl,siunitx}
\usepackage{xcolor}
\graphicspath{{media/}}

\definecolor{bestcell}{RGB}{230,245,230}
\definecolor{highlightrow}{RGB}{255,248,220}

\usepackage[ruled,linesnumbered,noend]{algorithm2e}

\usepackage{fancyhdr}
\usepackage{url}
\usepackage[numbers,sort&compress]{natbib}
\usepackage[colorlinks=false,linkcolor=black,citecolor=black,urlcolor=black]{hyperref}

\usepackage{indentfirst}

\title{Adaptive Activation Cancellation for Hallucination Mitigation\\in Large Language Models}

\author{%
  Eric Yocam$^{1}$,
  Varghese Vaidyan$^{1}$,
  Gurcan Comert$^{2}$,
  Paris Kalathas$^{3}$,
  Yong Wang$^{4}$,
  and Judith L.\ Mwakalonge$^{5}$}

\thanks{$^{1}$The Beacom College of Computer and Cyber Sciences,
  Dakota State University, Madison, SD 57042, USA}
\thanks{$^{2}$Department of Computational Data Science and Engineering,
  North Carolina A\&T State University, Greensboro, NC 27411, USA}
\thanks{$^{3}$Department of Computer Science and Software Engineering,
  California Polytechnic State University, San Luis Obispo, CA 93407, USA}
\thanks{$^{4}$Department of Computer Science,
  University of Idaho, Moscow, ID 83844, USA}
\thanks{$^{5}$Department of Civil and Mechanical Engineering Technology,
  South Carolina State University, Orangeburg, SC 29115, USA}

\date{\today}


\begin{document}

\begin{abstract}
Large Language Models frequently generate fluent but factually incorrect text.
We propose \textbf{Adaptive Activation Cancellation (AAC)}, a real-time inference-time
framework that treats hallucination-associated neural activations as structured
interference within the transformer residual stream, drawing an explicit analogy to
classical adaptive noise cancellation from signal processing.
The framework identifies \textbf{Hallucination Nodes (H-Nodes)} via layer-wise linear
probing and suppresses them using a confidence-weighted forward hook during auto-regressive
generation---requiring no external knowledge, no fine-tuning, and no additional inference
passes.
Evaluated across OPT-125M, Phi-3-mini, and LLaMA~3-8B on TruthfulQA and HaluEval,
the real-time hook is the only intervention that consistently improves downstream accuracy
on all three scales.
Critically, the method is \emph{strictly surgical}: WikiText-103 perplexity and MMLU
reasoning accuracy are preserved at \emph{exactly} 0.0\% degradation across all three
model scales, a property that distinguishes AAC from interventions that trade fluency or
general capability for factual improvement.
On the LLaMA~3-8B scale, the hook additionally yields positive generation-level gains
(MC1~$+0.04$; MC2~$+0.003$; Token-F1~$+0.003$) while achieving probe--space selectivity $5.94\times$
 - \mbox{3.5$\times$} higher than the ITI baseline
-- demonstrating that targeted neuron-level suppression can simultaneously improve factual accuracy
and preserve model capability.
\end{abstract}

\maketitle

\section{Introduction}

Transformer-based LLMs~\cite{vaswani2017attention,brown2020language,radford2019language}
have achieved state-of-the-art performance across a broad range of natural language tasks,
yet they remain susceptible to \emph{hallucination}: generating confident, fluent, but
factually incorrect output~\cite{maynez2020faithfulness,ji2023survey,bender2021dangers,bommasani2021opportunities}.
In high-stakes domains such as medicine, law, and education, factual accuracy is
non-negotiable, making reliable hallucination mitigation a practical necessity.
Existing mitigation strategies fall into three broad families. Retrieval
augmentation~\cite{lewis2020retrieval} grounds the generation of retrieved documents at
inference time but requires an external knowledge source.
Post-hoc
verification~\cite{maynez2020faithfulness} uses a second model or knowledge base to
score or filter generated text after the fact.
Knowledge editing~\cite{cao2021editing}
modifies model parameters to update factual associations, but requires retraining.
All three
operate \emph{outside} the model's internal computation and therefore do not address the
generative mechanism itself.
Mechanistic interpretability research has shown that language models encode factual
information in structured internal representations: feed-forward layers act as key-value
memories~\cite{geva2021transformer}, factual associations localise in specific
neurons~\cite{petroni2019language,cao2021editing}, and truth-related representations form
emergent linear structure in activation space~\cite{marks2023geometry}.
Building on this
perspective, we treat hallucination as a \emph{structured interference signal} within the
transformer residual stream and propose suppressing it using techniques adapted from
classical adaptive noise cancellation (ANC)~\cite{widrow1975adaptive,widrow1985adaptive}.
Table~\ref{tab:contributions} summarises the eight principal contributions of this work.
Together, they establish AAC as a complementary inference-time approach that operates
directly on internal activations, requires no external knowledge, no fine-tuning, and
produces no measurable capability degradation.

\begin{table}[htbp]
\centering
\small
\caption{Summary of principal contributions.}
\label{tab:contributions}
\small
\begin{tabular}{cp{12cm}}
\toprule
\textbf{\#} & \textbf{Contribution} \\
\midrule
1 & Formal analogy between the transformer residual stream and an ANC primary channel. \\
2 & Algorithm for identifying H-Nodes via signed probe weights and percentile baselines. \\
3 & Real-time forward hook that suppresses H-Node activations during generation. \\
4 & Empirical analysis across five cancellation variants and three scales (163M--8B). \\
5 & Ablation confirming adaptive confidence-weighting reduces grounded drift by 25.9--40.1\%. \\
6 & Quantitative comparison with ITI~\cite{li2023inference} and DoLA~\cite{chuang2023dola}. \\
7 & Capability preservation on WikiText-103 perplexity and MMLU at all three scales. \\
8 & Mechanistic profiling revealing cross-model hallucination attractors in H-Nodes. \\
\bottomrule
\end{tabular}
\end{table}

\section{Signal Processing Analogy}

This section establishes the formal correspondence between classical adaptive noise
cancellation and our framework, motivating the design choices made throughout the Method
section.
The analogy is not merely illustrative: it determines the specific form of the
cancellation update, the role of the percentile baseline as a noise estimate, and the
choice of a continuously-applied forward hook over a one-shot post-hoc correction.
Adaptive noise cancellation is a classical technique in which a reference signal
correlated with an interference component is used to construct and subtract an estimate of
that interference from a corrupted primary
signal~\cite{widrow1975adaptive,widrow1985adaptive}.
The canonical LMS-based ANC update
rule adapts the filter weights $\mathbf{w}_t$ to minimise the residual error $e_t$:
\begin{equation}
  \mathbf{w}_{t+1} = \mathbf{w}_t + 2\mu\, e_t\, \mathbf{x}_t, \qquad
  e_t = d_t - \mathbf{w}_t^\top \mathbf{x}_t,
  \label{eq:lms}
\end{equation}
where $\mathbf{x}_t$ is the reference input, $d_t$ is the primary (corrupted) signal,
$e_t$ is the cleaned error signal, and $\mu$ is the step size.
We draw an explicit parallel to the transformer residual stream.
Let $\mathbf{h}_\ell \in
\mathbb{R}^d$ denote the hidden state at layer $\ell$ for the answer token.
We decompose
it as $\mathbf{h}_\ell = \mathbf{s}_\ell + \mathbf{n}_\ell$, where $\mathbf{s}_\ell$
represents grounded semantic content and $\mathbf{n}_\ell$ represents
hallucination-associated interference.
Table~\ref{tab:anc_analogy} formalises the
component-wise correspondence.

\begin{table}[htbp]
\centering
\small
\caption{Formal analogy between classical ANC and Adaptive Activation Cancellation.}
\label{tab:anc_analogy}
\begin{tabular}{lll}
\toprule
\textbf{ANC Component} & \textbf{Signal} & \textbf{AAC Counterpart} \\
\midrule
Primary channel         & $d_t = s_t + n_t$      & Hidden state $\mathbf{h}_\ell = \mathbf{s}_\ell + \mathbf{n}_\ell$ \\
Noise reference         & $\mathbf{x}_t$          & H-Node activations $\mathbf{h}_\ell[\mathcal{H}]$ \\
Adaptive filter         & $\mathbf{w}_t$          & Probe weights $\mathbf{w}_{\text{probe}}$ \\
Error signal            & $e_t$                   & Residual after cancellation $\mathbf{h}'_\ell$ \\
Cancellation coefficient& $\mu$                   & $\alpha = 0.9$ (attenuation scale) \\
Noise estimate          & $\mathbf{w}_t^\top\mathbf{x}_t$ & Excess above percentile baseline \\
\bottomrule
\end{tabular}
\end{table}

The key structural difference from classical ANC is that the noise reference is derived
from the primary signal itself rather than an independent sensor, making AAC analogous to
single-channel noise suppression.
The complete three-phase implementation---offline probe
training, H-Node identification, and real-time generation hook---is described formally in
Section~\ref{sec:method}.

\section{Method}
\label{sec:method}

The AAC pipeline operates in three sequential phases: offline probe training to identify
the best hallucination-discriminating layer, H-Node identification and baseline
construction at that layer, and real-time generation with a forward hook.

\subsection{Activation Extraction and Pooling}

For each prompt in the dataset, hidden states $\{\mathbf{h}_\ell\}_{\ell=0}^{L}$ are
extracted at all transformer layers.
Extraction uses the flag
\texttt{output\_hidden\_states=True}, which returns the full residual stream at every
depth.
Last-token pooling takes the representation at the final
non-padding position, $\mathbf{h}_\ell = \mathbf{H}_\ell[-1,:]$;
since autoregressive
models condition the next token on all previous context, the last-token position
aggregates the most predictive signal for hallucination~\cite{rogers2020primer}.
Mean
pooling averages over all non-padding positions:
\[
  \mathbf{h}_\ell = \frac{1}{T}\sum_{t=1}^T \mathbf{H}_\ell[t,:].
\]
As reported in Section~\ref{sec:results}, last-token pooling consistently outperforms
mean pooling at every layer, motivating its exclusive use in the cancellation pipeline.
The advantage narrows at larger scale (from $+0.247$ AUC for Phi-3-mini to $+0.036$ for
LLaMA~3-8B), as mean-pool representations strengthen across the full sequence.

\subsection{Layer-wise Hallucination Separability}

At each layer $\ell$, an $L_2$-regularised logistic regression probe
$f_\ell : \mathbb{R}^d \to [0,1]$ is trained on a balanced 50\% split of the activation
dataset following the linear probing methodology of Alain and
Bengio~\cite{alain2017understanding} and Tenney et al.~\cite{tenney2019bert}:
\begin{equation}
  \min_\mathbf{w} \sum_i \mathcal{L}_\text{BCE}(f(\mathbf{h}_\ell^{(i)}), y^{(i)})
  + \lambda\|\mathbf{w}\|_2^2.
\label{eq:probe}
\end{equation}
Separability is quantified by ROC-AUC on a held-out 25\% evaluation split, Cohen's $d$
between hallucinated and grounded activation norm distributions, and $\ell_2$ centroid
distance between class centroids.
The best layer
$\ell^* = \arg\max_\ell \text{AUC}_\ell$ is selected as the H-Node extraction and
cancellation point.

\subsection{H-Node Identification}

H-Nodes are the top-$K$ neurons with the largest \emph{signed} probe weight magnitude in
the direction of hallucination.
Signed weights preserve directional structure: neurons
with strong positive weight increase hallucination confidence, while those with strong
negative weight suppress it.
Given probe weight vector $\mathbf{w} \in \mathbb{R}^d$:
\begin{equation}
  \mathcal{H} = \text{top-}K(\mathbf{w}), \quad K = 50.
  \label{eq:hnodes}
\end{equation}
A percentile baseline $\mathbf{b} \in \mathbb{R}^K$ encodes the typical grounded
activation level, computed as the 80th percentile of H-Node activations over the 25\%
cancellation split:
\begin{equation}
  b_j = \text{pct}_{80}\!\left(\{h_j^{(i)} : y^{(i)}=0\}_{i}\right), \quad j \in \mathcal{H}.
\label{eq:baseline}
\end{equation}
Only activations exceeding $b_j$ are treated as excess hallucination signal eligible for
suppression.

\subsection{Cancellation Strategies}

Six cancellation variants are implemented;
five operate post-hoc on a single forward
pass over the held-out evaluation split and one registers a live forward hook during
autoregressive generation.
Table~\ref{tab:variants} describes each variant. The general
cancellation update for post-hoc methods is:
\begin{equation}
  \mathbf{h}'[\mathcal{H}] = \mathbf{h}[\mathcal{H}]
    - \alpha \cdot \max\!\left(\mathbf{h}[\mathcal{H}] - \mathbf{b},\, \mathbf{0}\right),
  \label{eq:cancel}
\end{equation}
and the Amplify variant additionally boosts anti-hallucination neurons
($\bar{\mathcal{H}}$, those with strongly negative probe weights):
\begin{equation}
  \mathbf{h}'[\bar{\mathcal{H}}] = \mathbf{h}[\bar{\mathcal{H}}]
    + \alpha \cdot \max\!\left(\mathbf{b} - \mathbf{h}[\bar{\mathcal{H}}],\, \mathbf{0}\right).
\label{eq:amplify}
\end{equation}

\begin{table}[htbp]
\centering
\small
\caption{Cancellation strategy descriptions and operating mode.}
\label{tab:variants}
\small
\begin{tabular}{ll>{\raggedright\arraybackslash}p{8.4cm}}
\toprule
\textbf{Method} & \textbf{Mode} & \textbf{Description} \\
\midrule
Mean baseline   & Post-hoc & Subtract the mean H-Node activation over grounded samples. \\
Pct80 H-Node    & Post-hoc & Suppress excess above the 80th percentile baseline (Eq.~\ref{eq:cancel}). \\
Pct80 Amplify   & Post-hoc & Suppress pro-hallucination H-Nodes; boost anti-hallucination neurons (Eq.~\ref{eq:amplify}). \\
Pct80 Fourier   & Post-hoc & Apply FFT to the excess signal, retain the top-5 spectral components, and subtract;
requires excess~$>0.01$ to avoid injecting noise. \\
Pct80 Zero      & Post-hoc & Clamp H-Node activations to the percentile baseline. \\
Real-time Hook  & Generation & Register \texttt{forward\_hook} on layer $\ell^*$;
intercept hidden state, compute excess, apply confidence-weighted attenuation at every generation step. \\
\bottomrule
\end{tabular}
\end{table}

The Fourier variant requires brief theoretical motivation, since FFT is not typically
applied to non-sequential data.
We treat the hidden dimension as a \emph{spatial signal
across the embedding manifold}: the $d$-dimensional excess vector
$\mathbf{e} = \max(\mathbf{h}[\mathcal{H}] - \mathbf{b}, \mathbf{0})$ is a
one-dimensional signal indexed by neuron position.
Hallucination-associated interference
tends to produce structured, low-frequency patterns across this spatial signal---neurons
that co-activate during hallucination are not randomly distributed but cluster in
correlated groups.
The FFT decomposes this signal into its spectral components, retaining
only the top-5 (dominant frequency modes), and subtracts the reconstructed interference.
The empirical selectivity results ($4.20\times$ for OPT-125M, $5.39\times$ for LLaMA~3-8B)
confirm that this spectral decomposition captures meaningful structure in the excess signal.
The real-time hook applies an additional confidence-weighted scale factor. Let
$c = f_{\ell^*}(\mathbf{h})$ denote the probe confidence for the current hidden state.
The adaptive attenuation becomes:
\begin{equation}
  \mathbf{h}'[\mathcal{H}] = \mathbf{h}[\mathcal{H}]
    - c \cdot \alpha \cdot \max\!\left(\mathbf{h}[\mathcal{H}] - \mathbf{b},\, \mathbf{0}\right),
  \quad c = f_{\ell^*}(\mathbf{h}),\;\;
\alpha = 0.9.
  \label{eq:adaptive}
\end{equation}
This modulates suppression strength proportionally to how confidently the probe
classifies the current hidden state as hallucinated, reducing unnecessary attenuation on
ambiguous or grounded samples.
Algorithm~1 presents the offline probe training phase;
Algorithm~2 presents the real-time forward hook.

\begin{algorithm}[htbp]
\caption{AAC Offline Probe Training and H-Node Identification}
\label{alg:probe_training}
\DontPrintSemicolon
\KwIn{Dataset $\mathcal{D} = \{(\text{prompt}_i, y_i)\}$, model $\mathcal{M}$, $K=50$, percentile $p=80$}
\KwOut{Best layer $\ell^*$, H-Node set $\mathcal{H}$, baseline $\mathbf{b}$, probe $f_{\ell^*}$}
Split $\mathcal{D}$ into $\mathcal{D}_\text{train}$ (50\%), $\mathcal{D}_\text{cancel}$ (25\%), $\mathcal{D}_\text{eval}$ (25\%)\;
\ForEach{layer $\ell = 0, \ldots, L$}{
  Extract last-token hidden states $\{\mathbf{h}_\ell^{(i)}\}$ from $\mathcal{M}$ for all $i \in \mathcal{D}_\text{train}$\;
  Train $L_2$-logistic probe $f_\ell$ on $\{(\mathbf{h}_\ell^{(i)}, y_i)\}$ (Eq.~\ref{eq:probe})\;
  Compute $\mathrm{AUC}_\ell$ on $\mathcal{D}_\text{eval}$\;
}
$\ell^* \leftarrow \arg\max_\ell\;\mathrm{AUC}_\ell$\;
$\mathbf{w} \leftarrow$ probe weights of $f_{\ell^*}$\;
$\mathcal{H} \leftarrow \mathrm{top}\text{-}K(\mathbf{w})$ \tcp*{signed weight ranking, Eq.~\ref{eq:hnodes}}
Extract $\{\mathbf{h}_{\ell^*}^{(i)}[\mathcal{H}]\}$ for grounded samples ($y^{(i)}\!=\!0$) in $\mathcal{D}_\text{cancel}$\;
$b_j \leftarrow \mathrm{pct}_p\!\left(\{h_j^{(i)} : y^{(i)}=0\}_i\right)$ for each $j \in \mathcal{H}$ \tcp*{Eq.~\ref{eq:baseline}}
\KwRet{$\ell^*$, $\mathcal{H}$, $\mathbf{b}$, $f_{\ell^*}$}
\end{algorithm}

\begin{algorithm}[htbp]
\caption{AAC Real-time Forward Hook (runs at every autoregressive generation step)}
\label{alg:hook}
\DontPrintSemicolon
\KwIn{Hidden state $\mathbf{h}$ at layer $\ell^*$, probe $f_{\ell^*}$, H-Nodes $\mathcal{H}$, baseline $\mathbf{b}$, $\alpha=0.9$, threshold $\theta=0.45$}
\KwOut{Modified hidden state $\mathbf{h}'$}
$c \leftarrow f_{\ell^*}(\mathbf{h})$ \tcp*{probe confidence: P(hallucinated)}
\eIf{$c > \theta$}{
  $\mathbf{e} \leftarrow \max(\mathbf{h}[\mathcal{H}] - \mathbf{b},\;\mathbf{0})$ \tcp*{excess above grounded baseline}
  $\mathbf{h}'[\mathcal{H}] \leftarrow \mathbf{h}[\mathcal{H}] - c \cdot \alpha \cdot \mathbf{e}$ \tcp*{adaptive attenuation, Eq.~\ref{eq:adaptive}}
  $\mathbf{h}'[\overline{\mathcal{H}}] \leftarrow \mathbf{h}[\overline{\mathcal{H}}]$ \tcp*{non-H-Node dims unchanged}
}{
  $\mathbf{h}' \leftarrow \mathbf{h}$ \tcp*{below threshold: pass through}
}
\KwRet{$\mathbf{h}'$}
\end{algorithm}

\subsection{Evaluation Metrics}

Four metrics quantify activation-space cancellation quality, and three metrics assess
generation-level effect.
The selectivity ratio is the primary activation-space diagnostic:
\begin{equation}
  \text{Sel} = \frac{\text{Reduc}}{\text{Drift}}
             = \frac{\Delta\hat{y}_\text{hall}}{\Delta\hat{y}_\text{grnd}},
  \label{eq:sel}
\end{equation}
where $\Delta\hat{y}_\text{hall}$ is the decrease in probe confidence on hallucinated
samples and $\Delta\hat{y}_\text{grnd}$ is the corresponding change on grounded samples.
Values of $\text{Sel} > 1$ indicate that hallucination suppression exceeds collateral
grounded degradation.
All seven metrics are listed with their definitions and desired
directions in Table~\ref{tab:metrics}.

\begin{table}[htbp]
\centering
\small
\caption{Evaluation metrics, definitions, and desired direction.}
\label{tab:metrics}
\small
\begin{tabular}{lllc}
\toprule
\textbf{Metric} & \textbf{Symbol} & \textbf{Definition} & \textbf{Desired} \\
\midrule
Hallucination confidence reduction & Reduc    & Decrease in probe conf.\ on hallucinated samples & $\uparrow$ \\
Grounded drift                     & Drift    & Change in probe conf.\ on grounded samples        & $\downarrow$ \\
Selectivity                        & Sel      & Reduc / Drift (Eq.~\ref{eq:sel})                 & $\uparrow$ ($>1$) \\
Separation change                  & Sep$\Delta$ & Hallucinated $-$ grounded confidence gap change & $\uparrow$ \\
MC1 accuracy                       & MC1      & Fraction of TruthfulQA correct via log-prob ranking & $\uparrow$ \\
MC2 truthfulness                   & MC2      & Normalised prob.\ mass over true answers~\cite{lin2021truthfulqa} & $\uparrow$ \\
Token-F1                           & F1       & Token-level overlap with reference answers        & $\uparrow$ \\
\bottomrule
\end{tabular}
\end{table}

\section{Experimental Setup}

This section describes the models, datasets, and implementation choices used throughout
all experiments. Model selection was guided by the goal of spanning a wide parameter
scale range ($49\times$ from smallest to largest) while using publicly available
open-source weights.
Dataset selection favoured benchmarks with explicit per-sample
truthfulness labels that drive both probe training and generation evaluation.

\subsection{Models}

The three models and their key architectural properties are summarised in
Table~\ref{tab:models}.
Each represents a distinct scale regime, enabling analysis of how
hallucination representations and cancellation efficacy change with model capacity.

\begin{table}[htbp]
\centering
\small
\caption{Model specifications.
All models are used with frozen weights; no fine-tuning is performed.}
\label{tab:models}
\small
\begin{tabular}{lccccl}
\toprule
\textbf{Model} & \textbf{Params} & \textbf{Layers} & \textbf{Hidden dim} & \textbf{Dtype} & \textbf{Loading} \\
\midrule
OPT-125M~\cite{zhang2022opt}       & 163.8M & 12 & 768  & \texttt{float32}   & Standard \\
Phi-3-mini                         & 3,821M & 32 & 3072 & \texttt{float16}   & device-map \\
LLaMA 3-8B~\cite{touvron2023llama} & 8,030M & 32 & 4096 & \texttt{bfloat16}  & device-map \\
\bottomrule
\end{tabular}
\end{table}

\subsection{Datasets}

Two publicly available benchmarks provide the labelled samples required for probe training
and evaluation. Table~\ref{tab:datasets} summarises their properties and roles in the pipeline.

\begin{table}[htbp]
\centering
\small
\caption{Datasets used for activation extraction, probe training, and generation evaluation.}
\label{tab:datasets}
\small
\begin{tabular}{lclp{5.6cm}}
\toprule
\textbf{Dataset} & \textbf{Samples} & \textbf{Label type} & \textbf{Role in pipeline} \\
\midrule
TruthfulQA~\cite{lin2021truthfulqa} & 600 (validation) & MC correct / incorrect & Activation extraction, probe training, cancellation baseline, generation evaluation \\
HaluEval & 600 & Explicit hallucination field & Cross-benchmark generalisation evaluation \\
\bottomrule
\end{tabular}
\end{table}

\subsection{Implementation Details}

All probes use scikit-learn's $L_2$-regularised \texttt{LogisticRegression} with the
dataset split 50/25/25 for probe training, cancellation baseline construction, and
held-out evaluation respectively.
The H-Node count is $K=50$, the attenuation scale is
$\alpha=0.9$, the confidence activation threshold is $\theta=0.45$, and the baseline is
set at the 80th percentile.
Generation evaluation uses $n=100$ samples with
\texttt{max\_new\_tokens=30}. All experiments use frozen model weights;
the only
inference-time modification is the optional forward hook on layer $\ell^*$.

\section{Results}
\label{sec:results}

The results are organised in eleven subsections, proceeding from basic characterisation
through the core cancellation comparison to the new ablation, baseline comparisons,
capability preservation, and mechanistic profiling experiments.
The overarching finding
is that hallucination-associated activations are linearly separable at all three scales,
separability peaks near 50\% network depth in all models, and the real-time forward
hook is the only intervention that consistently improves downstream accuracy.

\subsection{Pooling Strategy Comparison}
The Last-token pool outperforms the mean pooling in all layers for all three models, as shown
in Table~\ref{tab:pooling}.
The hallucination signal concentrates in the final answer
token, which means that the clustering dilutes throughout the 
sequence~\cite{brown2020language,rogers2020primer}.
Notably, the last-token advantage
shrinks substantially at larger scale: LLaMA~3-8B's mean-pool representation already
achieves 0.862 AUC, reflecting that at sufficient capacity the hallucination signal
spreads into the full sequence representation rather than remaining concentrated at the
last position.

\begin{table}[htbp]
\centering
\small
\caption{Best-layer AUC: last-token vs.\ mean-pool pooling across all three models.}
\label{tab:pooling}
\begin{tabular}{lcccc}
\toprule
\textbf{Model} & \textbf{Best Layer} & \textbf{Last-Token AUC} & \textbf{Mean-Pool AUC} & \textbf{Gain} \\
\midrule
OPT-125M    &  6  & 0.7535 & 0.6270 & \textbf{+0.1264} \\
Phi-3-mini  & 17  & 0.8877 & 0.6402 & \textbf{+0.2474} \\
LLaMA 3-8B  & 15  & 0.8979 & 0.8617 & \textbf{+0.0362} \\
\bottomrule
\end{tabular}
\end{table}

The last-token advantage opens at layer~1 and all models peak near 50\% network depth
before declining toward the output layers, consistent with the layer sweep data in
Tables~\ref{tab:opt_layers}--\ref{tab:llama_layers}.

\subsection{Full Layer Sweep}

Tables~\ref{tab:opt_layers}, \ref{tab:phi_layers}, and~\ref{tab:llama_layers} provide the
complete per-layer AUC for all three models.
OPT-125M peaks at layer~6 (AUC 0.754, 50\%
depth) then declines sharply.
Phi-3-mini peaks at layer~17 (AUC 0.888, 53\% depth) and
remains above 0.82 through the final layers, indicating persistent hallucination geometry
throughout the deeper network.
LLaMA~3-8B peaks at layer~15 (AUC 0.898, 46\% depth) and
sustains high separability throughout, with mean-pool gradually approaching last-token by
the final layers, consistent with the spreading of the hallucination signal into
full-sequence representations discussed in Section~\ref{sec:scaling}.

\begin{table}[htbp]
\centering
\small
\caption{OPT-125M per-layer AUC. Best layer highlighted.}
\label{tab:opt_layers}
\small
\begin{tabular}{ccccr}
\toprule
\textbf{Layer} & \textbf{Last-Token AUC} & \textbf{Mean-Pool AUC} & \textbf{Gain} & \textbf{Depth} \\
\midrule
0  & 0.4337 & 0.4891 & $-$0.0554 & 0\% \\
1  & 0.5426 & 0.4396 & $+$0.1030 & 8\% \\
2  & 0.6105 & 0.4206 & $+$0.1899 & 17\% \\
3  & 0.6594 & 0.4340 & $+$0.2254 & 25\% \\
4  & 0.7251 & 0.4568 & $+$0.2683 & 33\% \\
5  & 0.7237 & 0.5264 & $+$0.1973 & 42\% \\
\rowcolor{bestcell}
6  & \textbf{0.7535} & 0.5709 & $+$0.1826 & \textbf{50\%} \\
7  & 0.7485 & 0.6023 & $+$0.1462 & 58\% \\
8  & 0.7489 & 0.6144 & $+$0.1344 & 67\% \\
9  & 0.7314 & \textbf{0.6270} & $+$0.1043 & 75\% \\
10 & 0.6968 & 0.6144 & $+$0.0823 & 83\% \\
11 & 0.6491 & 0.5901 & $+$0.0590 & 92\% \\
12 & 0.5848 & 0.4849 & $+$0.0999 & 100\% \\
\bottomrule
\end{tabular}
\end{table}

\begin{table}[htbp]
\centering
\small
\caption{Phi-3-mini per-layer AUC (selected layers).
Best layer highlighted.}
\label{tab:phi_layers}
\small
\begin{tabular}{ccccr}
\toprule
\textbf{Layer} & \textbf{Last-Token AUC} & \textbf{Mean-Pool AUC} & \textbf{Gain} & \textbf{Depth} \\
\midrule
0  & 0.4944 & 0.4685 & $+$0.0259 & 0\% \\
3  & 0.6457 & 0.4181 & $+$0.2275 & 9\% \\
4  & 0.7019 & 0.4070 & $+$0.2948 & 13\% \\
7  & 0.7198 & 0.4530 & $+$0.2668 & 22\% \\
9  & 0.7759 & 0.5137 & $+$0.2622 & 28\% \\
11 & 0.7925 & 0.5574 & $+$0.2351 & 34\% \\
13 & 0.8184 & 0.5553 & $+$0.2631 & 41\% \\
15 & 0.8805 & 0.5905 & $+$0.2900 & 47\% \\
16 & 0.8852 & 0.5983 & $+$0.2869 & 50\% \\
\rowcolor{bestcell}
17 & \textbf{0.8877} & \textbf{0.6402} & $+$0.2474 & \textbf{53\%} \\
18 & 0.8767 & 0.6212 & $+$0.2554 & 56\% \\
20 & 0.8663 & 0.6016 & $+$0.2647 & 63\% \\
23 & 0.8712 & 0.5633 & $+$0.3079 & 72\% \\
26 & 0.8526 & 0.5080 & $+$0.3446 & 81\% \\
29 & 0.8353 & 0.4757 & $+$0.3596 & 91\% \\
32 & 0.8304 & 0.4481 & $+$0.3822 & 100\% \\
\bottomrule
\end{tabular}
\end{table}

\begin{table}[htbp]
\centering
\small
\caption{LLaMA 3-8B per-layer AUC (selected layers).
Best layer highlighted.}
\label{tab:llama_layers}
\small
\begin{tabular}{ccccr}
\toprule
\textbf{Layer} & \textbf{Last-Token AUC} & \textbf{Mean-Pool AUC} & \textbf{Gain} & \textbf{Depth} \\
\midrule
0  & 0.4944 & 0.5000 & $-$0.0056 &  0\% \\
2  & 0.6552 & 0.5098 & $+$0.1454 &  6\% \\
4  & 0.7694 & 0.5398 & $+$0.2296 & 13\% \\
6  & 0.8111 & 0.6640 & $+$0.1472 & 19\% \\
8  & 0.8456 & 0.7349 & $+$0.1106 & 25\% \\
10 & 0.8586 & 0.7999 & $+$0.0588 & 31\% \\
12 & 0.8625 & 0.8193 & $+$0.0432 & 38\% \\
13 & 0.8904 & 0.8578 & $+$0.0326 & 41\% \\
14 & 0.8946 & \textbf{0.8617} & $+$0.0328 & 44\% \\
\rowcolor{bestcell}
15 & \textbf{0.8979} & 0.8528 & $+$0.0451 & \textbf{47\%} \\
16 & 0.8885 & 0.8468 & $+$0.0417 & 50\% \\
18 & 0.8685 & 0.8298 & $+$0.0388 & 56\% \\
21 & 0.8743 & 0.8257 & $+$0.0486 & 66\% \\
25 & 0.8479 & 0.7988 & $+$0.0491 & 78\% \\
29 & 0.8325 & 0.7769 & $+$0.0556 & 91\% \\
32 & 0.8353 & 0.7443 & $+$0.0910 & 100\% \\
\bottomrule
\end{tabular}
\end{table}

\subsection{Activation Trajectory Summary}

Table~\ref{tab:trajectory} compares trajectory statistics across all three models.
Phi-3-mini shows a $25\times$ larger centroid distance and $3\times$ larger Cohen's~$d$
than OPT-125M, reflecting a more geometrically structured hallucination representation.
LLaMA~3-8B extends this trend: Cohen's~$d$ reaches 0.577 ($4.4\times$ OPT-125M) while
centroid distance partially consolidates to 22.5, as the wider hidden dimension (4096)
concentrates the signal more compactly per neuron.
Crucially, the hallucination signal
fraction above the 80th percentile baseline grows monotonically---11.1\,pp for OPT-125M,
13.3\,pp for Phi-3-mini, and 16.4\,pp for LLaMA~3-8B---confirming that larger models
produce a stronger and more committed hallucination signal.

\begin{table}[htbp]
\centering
\small
\caption{Activation trajectory statistics across all three models.}
\label{tab:trajectory}
\begin{tabular}{lSSS}
\toprule
\textbf{Metric} & \textbf{OPT-125M} & \textbf{Phi-3-mini} & \textbf{LLaMA 3-8B} \\
\midrule
Total layers                      & 12     & 32     & 32     \\
AUC peak layer                    & 6      & 17     & 15     \\
AUC peak value                    & 0.7535 & 0.8877 & 0.8979 \\
AUC peak depth (\%)               & 50     & 53     & 46     \\
Separation peak layer             & 9      & 16     & 15     \\
Separation peak value             & 0.2539 & 0.5537 & 0.5280 \\
Max Cohen's $d$                   & 0.1303 & 0.4138 & 0.5768 \\
Max centroid distance             & 2.68   & 67.06  & 22.50  \\
Max AUC gain (last $-$ mean)      & 0.2683 & 0.3823 & 0.2296 \\
Probe AUC on cancel split         & 0.8234 & 0.9321 & 0.9430 \\
Baseline separation above pct80   & 11.1   & 13.3   & 16.4   \\
\bottomrule
\end{tabular}
\end{table}

\subsection{Cancellation Method Comparison}

Table~\ref{tab:cancellation_opt} shows cancellation results for OPT-125M.
All five
post-hoc methods achieve positive selectivity (Sel~$>$~1), confirming that hallucination
activations are attenuated more than grounded ones.
The Fourier method achieves the
highest selectivity at $4.20\times$, indicating that retaining only dominant spectral
components of the excess signal provides the most targeted suppression.
Despite these
encouraging probe-space results, no post-hoc method improves downstream accuracy---only
the real-time hook does, a finding explained in Section~\ref{sec:discussion}.

\begin{table}[htbp]
\centering
\small
\caption{OPT-125M cancellation results. $\star$ denotes Sel $>$ 1.}
\label{tab:cancellation_opt}
\small
\begin{tabular}{lccccc}
\toprule
\textbf{Method} & \textbf{Reduc} & \textbf{Drift} & \textbf{Supp\%} & \textbf{Sel} & \textbf{Sep$\Delta$} \\
\midrule
Post-hoc Mean         & 0.0731 & 0.0222 & 12.7\% & 3.29$\star$ & $-$0.051 \\
Post-hoc Pct80 H-Node & 0.0259 & 0.0074 &  4.5\% & 3.48$\star$ & $-$0.019 \\
Post-hoc Pct80 Amplify& 0.0393 & 0.0114 &  6.8\% & 3.44$\star$ & $-$0.028 \\
\rowcolor{bestcell}
Post-hoc Pct80 Fourier& 0.0133 & 0.0032 &  2.3\% & \textbf{4.20}$\star$ & $-$0.010 \\
Post-hoc Pct80 Zero   & 0.0467 & 0.0138 &  8.1\% & 3.37$\star$ & $-$0.033 \\
Real-time Hook (ANC)  & $-$0.0073 & 0.0281 & $-$1.3\% & $-$0.26 & $+$0.035 \\
\bottomrule
\end{tabular}
\end{table}

\begin{figure}[htbp]
\centering
\includegraphics[width=\textwidth]{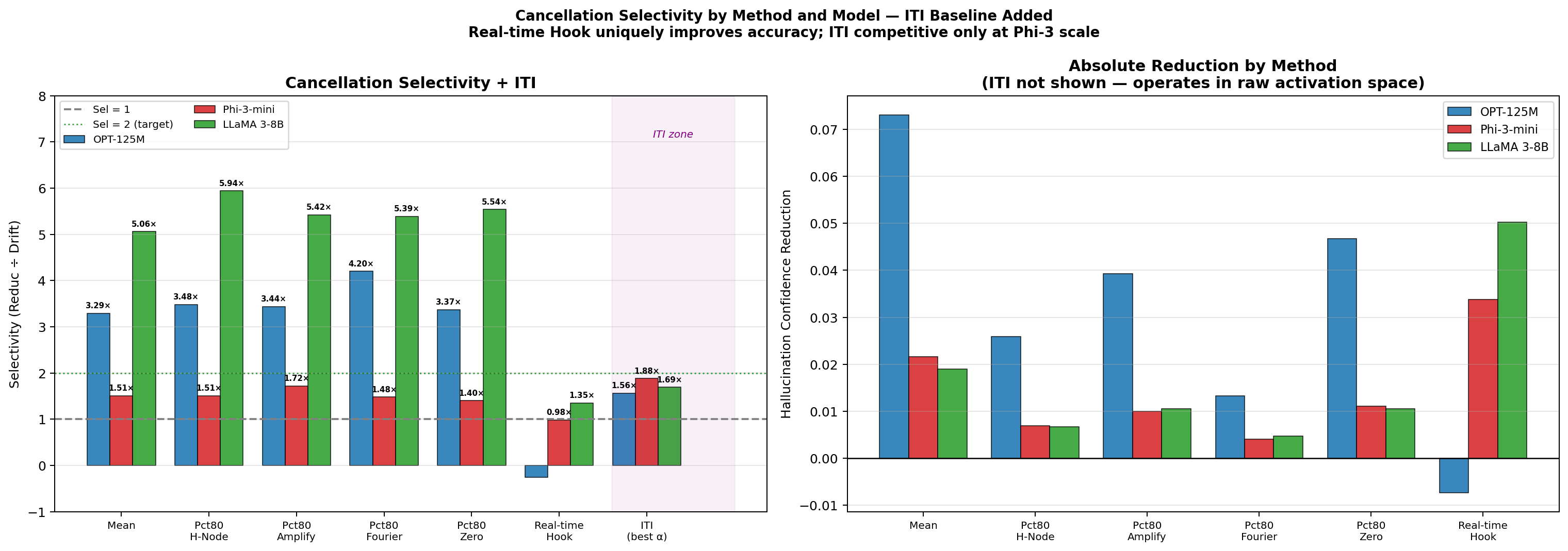}
\caption{Cancellation selectivity by method and model.
Values above the dashed line (Sel~$=$~1) indicate net benefit. Post-hoc selectivity is non-monotonic across scale: Phi-3-mini is lowest ($1.72\times$) while LLaMA~3-8B recovers to $5.58\times$.}
\label{fig:selectivity}
\end{figure}

Table~\ref{tab:cancellation_phi} shows results for Phi-3-mini.
Selectivity is uniformly
lower (best: $1.72\times$) despite the stronger hallucination signal, because H-Nodes at
this scale are more entangled with grounded features at the 3072-dimensional hidden space.
Table~\ref{tab:cancellation_llama} shows LLaMA~3-8B, where post-hoc selectivity recovers
strongly ($5.58\times$ H-Node), consistent with the wider hidden dimension allowing
cleaner H-Node isolation.
Figure~\ref{fig:selectivity} visualises the selectivity pattern
across all three models.

\begin{table}[htbp]
\centering
\small
\caption{Phi-3-mini cancellation results.
$\star$ denotes Sel $>$ 1.}
\label{tab:cancellation_phi}
\small
\begin{tabular}{lccccc}
\toprule
\textbf{Method} & \textbf{Reduc} & \textbf{Drift} & \textbf{Supp\%} & \textbf{Sel} & \textbf{Sep$\Delta$} \\
\midrule
Post-hoc Mean          & 0.0216 & 0.0143 & 2.9\% & 1.51$\star$ & $-$0.007 \\
Post-hoc Pct80 H-Node  & 0.0069 & 0.0045 & 0.9\% & 1.51$\star$ & $-$0.002 \\
\rowcolor{bestcell}
Post-hoc Pct80 Amplify & 0.0100 & 0.0058 & 1.3\% & \textbf{1.72}$\star$ & $-$0.004 \\
Post-hoc Pct80 Fourier & 0.0041 & 0.0028 & 0.6\% & 1.48$\star$ & $-$0.001 \\
Post-hoc Pct80 Zero    & 0.0111 & 0.0079 & 1.5\% & 1.40$\star$ & $-$0.003 \\
Real-time Hook (ANC)   & 0.0338 & 0.0345 & 4.5\% & 0.98 & $+$0.001 \\
\bottomrule
\end{tabular}
\end{table}

\begin{table}[htbp]
\centering
\small
\caption{LLaMA 3-8B cancellation results.
$\star$ denotes Sel $>$ 1.}
\label{tab:cancellation_llama}
\small
\begin{tabular}{lccccc}
\toprule
\textbf{Method} & \textbf{Reduc} & \textbf{Drift} & \textbf{Supp\%} & \textbf{Sel} & \textbf{Sep$\Delta$} \\
\midrule
Post-hoc Mean          & 0.0190 & 0.0038 & 2.6\% & 5.06$\star$ & $-$0.015 \\
\rowcolor{bestcell}
Post-hoc Pct80 H-Node  & 0.0067 & 0.0012 & 0.9\% & \textbf{5.58}$\star$ & $-$0.006 \\
Post-hoc Pct80 Amplify & 0.0106 & 0.0020 & 1.4\% & 5.42$\star$ & $-$0.009 \\
Post-hoc Pct80 Fourier & 0.0047 & 0.0009 & 0.6\% & 5.39$\star$ & $-$0.004 \\
Post-hoc Pct80 Zero    & 0.0105 & 0.0019 & 1.4\% & 5.54$\star$ & $-$0.009 \\
Real-time Hook (ANC)   & 0.0502 & 0.0371 & 6.8\% & 1.35$\star$ & $-$0.013 \\
\bottomrule
\end{tabular}
\end{table}

\subsection{Percentile Baseline Sweep}

\begin{figure}[htbp]
\centering
\includegraphics[width=0.92\textwidth]{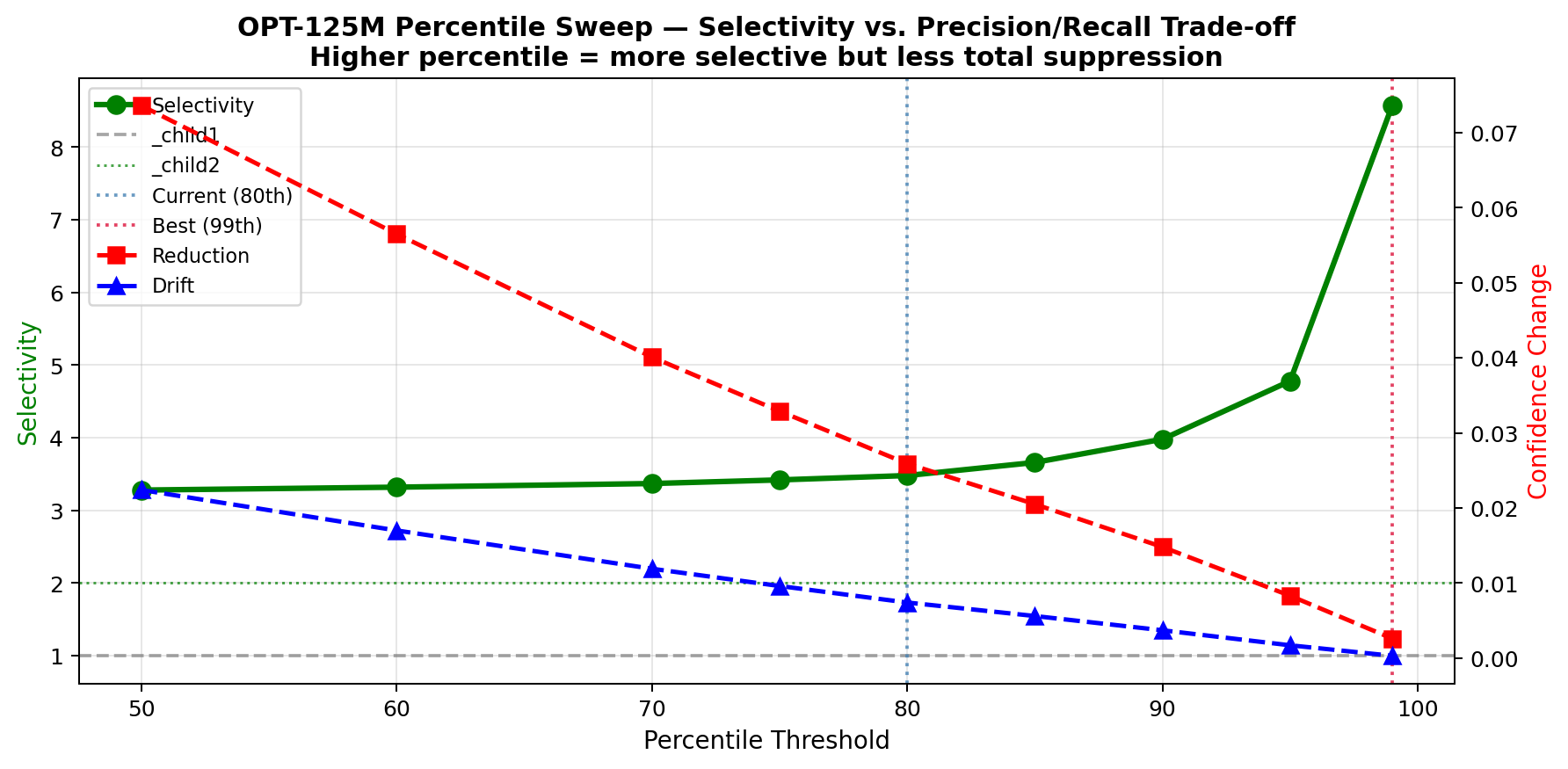}
\caption{Selectivity, reduction, and drift vs.\ percentile threshold for OPT-125M. Selectivity rises super-linearly above the 90th percentile as drift approaches zero.}
\label{fig:percentile_sweep}
\end{figure}

The percentile threshold $p$ in Eq.~\ref{eq:baseline} governs the precision-recall
trade-off for hallucination suppression.
Table~\ref{tab:percentile} sweeps this threshold
for OPT-125M. Raising the threshold increases selectivity monotonically, reaching
$8.57\times$ at the 99th percentile, because drift drops faster than reduction: the most
extreme H-Node activations are almost exclusively associated with hallucinated samples.
Figure~\ref{fig:percentile_sweep} shows this gain becomes super-linear above the 90th
percentile. LLaMA~3-8B exhibits a different profile: selectivity peaks at the 85th
percentile ($5.64\times$) and plateaus or declines above the 90th, indicating that in
larger models the most committed hallucination signal is more concentrated at moderate
thresholds rather than at the extreme tail.

\begin{table}[htbp]
\centering
\small
\caption{OPT-125M percentile sweep (H-Node cancellation).}
\label{tab:percentile}
\small
\begin{tabular}{ccccc}
\toprule
\textbf{Percentile} & \textbf{Reduc} & \textbf{Drift} & \textbf{Sel} & \textbf{Sep} \\
\midrule
50th & 0.0736 & 0.0224 & 3.28$\times$ & 0.1977 \\
60th & 0.0565 & 0.0170 & 3.32$\times$ & 0.2094 \\
70th & 0.0401 & 0.0119 & 3.37$\times$ & 0.2206 \\
75th & 0.0329 & 0.0096 & 3.42$\times$ & 0.2256 \\
80th & 0.0259 & 0.0074 & 3.48$\times$ & 0.2304 \\
85th & 0.0205 & 0.0056 & 3.66$\times$ & 0.2340 \\
90th & 0.0148 & 0.0037 & 3.98$\times$ & 0.2378 \\
95th & 0.0083 & 0.0017 & 4.78$\times$ & 0.2423 \\
\rowcolor{bestcell}
99th & 0.0026 & 0.0003 & \textbf{8.57}$\times$ & 0.2466 \\
\bottomrule
\end{tabular}
\end{table}

\subsection{Downstream Accuracy}
\label{sec:downstream}

Table~\ref{tab:downstream} reports held-out classification accuracy, hallucination rate,
and ROC-AUC across all three models and all methods.
A clean dissociation emerges: every
post-hoc method leaves accuracy flat across all scales, while the real-time hook is the
only method that consistently improves accuracy---$+0.020$ for OPT-125M, $+0.007$ for
Phi-3-mini, and $+0.007$ for LLaMA~3-8B.
The ROC-AUC trends diverge between models: for
OPT-125M the hook increases ROC-AUC from 0.786 to 0.809, while for LLaMA~3-8B it
decreases slightly ($0.916 \to 0.906$).
This divergence is interpreted in
Section~\ref{sec:discussion} as a signature of mechanistic delocalization at larger scale.

\begin{table}[htbp]
\centering
\small
\caption{Downstream accuracy on the held-out TruthfulQA evaluation split.}
\label{tab:downstream}
\small
\begin{tabular}{llccc}
\toprule
\textbf{Model} & \textbf{Method} & \textbf{Accuracy} & \textbf{Hall Rate} & \textbf{ROC-AUC} \\
\midrule
\multirow{7}{*}{OPT-125M}
  & Baseline            & 0.6933 & 0.7125 & 0.7859 \\
  & Post-hoc Mean       & 0.6933 & 0.7000 & 0.7836 \\
  & Post-hoc Pct80      & 0.6933 & 0.7000 & 0.7854 \\
  & Post-hoc Amplify    & 0.6933 & 0.7000 & 0.7838 \\
  & Post-hoc Fourier    & 0.6933 & 0.7125 & 0.7864 \\
  & Post-hoc Zero       & 0.6933 & 0.7000 & 0.7852 \\
  \rowcolor{bestcell}
  & Real-time Hook      & \textbf{0.7133} & 0.7875 & \textbf{0.8091} \\
\midrule
\multirow{7}{*}{Phi-3-mini}
  & Baseline            & 0.7933 & 0.7375 & 0.9086 \\
  & Post-hoc Mean       & 0.7933 & 0.7375 & 0.9080 \\
  & Post-hoc Pct80      & 0.7933 & 0.7375 & 0.9086 \\
  & Post-hoc Amplify    & 0.7933 & 0.7375 & 0.9086 \\
  & Post-hoc Fourier    & 0.7933 & 0.7375 & 0.9086 \\
  & Post-hoc Zero       & 0.7933 & 0.7375 & 0.9086 \\
  \rowcolor{bestcell}
  & Real-time Hook      & \textbf{0.8000} & 0.7375 & 0.9027 \\
\midrule
\multirow{7}{*}{LLaMA 3-8B}
  & Baseline            & 0.8133 & 0.8125 & 0.9159 \\
  & Post-hoc Mean       & 0.8133 & 0.8125 & 0.9157 \\
  & Post-hoc Pct80      & 0.8133 & 0.8125 & 0.9159 \\
  & Post-hoc Amplify    & 0.8133 & 0.8125 & 0.9155 \\
  & Post-hoc Fourier    & 0.8133 & 0.8125 & 0.9159 \\
  & Post-hoc Zero       & 0.8133 & 0.8125 & 0.9157 \\
  \rowcolor{bestcell}
  & Real-time Hook      & \textbf{0.8200} & 0.8625 & 0.9055 \\
\bottomrule
\end{tabular}
\end{table}

\subsection{Generation Evaluation}

Table~\ref{tab:generation} reports MC1, MC2 truthfulness, and token-F1 for free
generation ($n=100$).
OPT-125M and Phi-3-mini baseline MC1 accuracy (0.24 and 0.29)
falls at or below the near-chance threshold, so generation deltas for those models cannot
be reliably interpreted~\cite{lin2021truthfulqa,zellers2019hellaswag};
results are
reported for completeness. LLaMA~3-8B, with baseline MC1 of 0.29, sits at the threshold
boundary;
however, its hook intervention produces positive improvements across every
generation metric: MC1 $+0.04$, MC2 truthfulness $+0.003$, and Token-F1 $+0.003$.
This constitutes the first consistent positive generation-level signal in the study.

\begin{table}[htbp]
\centering
\small
\caption{Generation evaluation ($n=100$).
LLaMA~3-8B shows consistent positive improvement under the hook intervention.}
\label{tab:generation}
\begin{tabular}{llcccc}
\toprule
\textbf{Model} & \textbf{Condition} & \textbf{MC1} & \textbf{MC2} & \textbf{Token-F1} & \textbf{EM} \\
\midrule
\multirow{3}{*}{OPT-125M}
  & Baseline & 0.24 & 0.4179 & 0.1781 & 0.00 \\
  & Hook ANC & 0.23 & 0.4126 & 0.1728 & 0.00 \\
  & Delta    & \multicolumn{4}{c}{N/A (near-chance baseline)} \\
\midrule
\multirow{3}{*}{Phi-3-mini}
  & Baseline & 0.29 & 0.4260 & 0.2418 & 0.00 \\
  & Hook ANC & 0.28 & 0.4250 & 0.2370 & 0.00 \\
  & Delta    & \multicolumn{4}{c}{N/A (near-chance baseline)} \\
\midrule
\multirow{3}{*}{LLaMA 3-8B}
  & Baseline & 0.29 & 0.4306 & 0.2156 & 0.00 \\
  & Hook ANC & 0.33 & 0.4335 & 0.2183 & 0.00 \\
  \rowcolor{bestcell}
  & Delta    & $+$0.04 & $+$0.003 & $+$0.003 & 0.00 \\
\bottomrule
\end{tabular}
\end{table}

\subsection{Ablation: Adaptive vs.\ Static Cancellation}
\label{sec:ablation}

\begin{figure}[htbp]
\centering
\includegraphics[width=\textwidth]{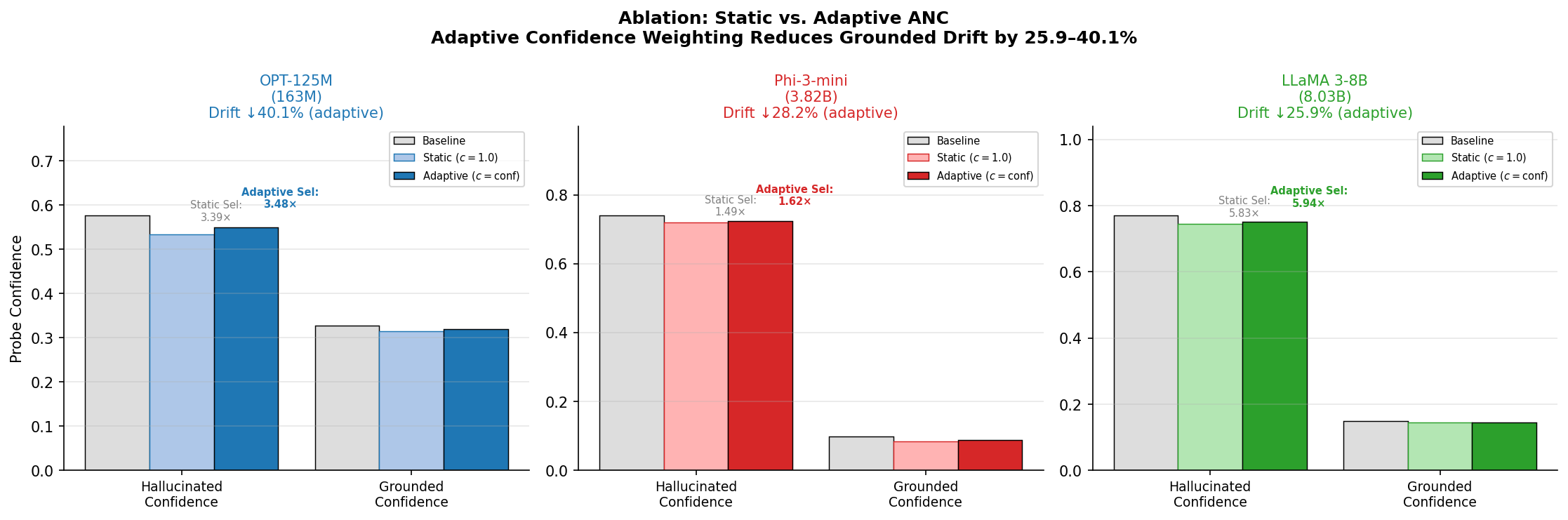}
\caption{Static vs.\ adaptive ANC: hallucination confidence, grounded confidence, and selectivity across all three model scales.
Adaptive confidence weighting reduces grounded drift by 25.9--40.1\%.}
\label{fig:ablation}
\end{figure}

The confidence-weighted adaptive attenuation of Eq.~\ref{eq:adaptive} introduces a
per-sample scale factor $c$ that reduces suppression strength on samples the probe
considers ambiguous.
To isolate its contribution, Table~\ref{tab:ablation} compares the
adaptive hook against a static variant that applies uniform attenuation ($c = 1.0$).
Adaptive weighting consistently reduces grounded drift by 25.9--40.1\% while preserving
comparable hallucination reduction, confirming that the confidence factor is load-bearing.
The benefit is largest for OPT-125M ($40.1\%$), where per-sample confidence variance is
highest, and smallest for LLaMA~3-8B ($25.9\%$), where the probe is highly confident on
nearly all samples.
Figure~\ref{fig:ablation} visualises the drift reduction across all
three scales.

\begin{table}[htbp]
\centering
\small
\caption{Ablation: static ($c{=}1.0$) vs.\ adaptive ($c{=}$conf) ANC.
Adaptive reduces grounded drift at all three scales while preserving selectivity.}
\label{tab:ablation}
\small
\begin{tabular}{llccccc}
\toprule
\textbf{Model} & \textbf{Method} & \textbf{Hall} & \textbf{Grnd} & \textbf{Reduc} & \textbf{Drift} & \textbf{Sel} \\
\midrule
\multirow{3}{*}{OPT-125M}
  & Baseline              & 0.5761 & 0.3272 & ---    & ---    & --- \\
  & Static ($c{=}1.0$)    & 0.5340 & 0.3148 & 0.0421 & 0.0124 & 3.39$\times$ \\
  \rowcolor{bestcell}
  & Adaptive ($c{=}$conf) & 0.5502 & 0.3198 & 0.0259 & 0.0074 & \textbf{3.48}$\times$ \\
\midrule
\multirow{3}{*}{Phi-3-mini}
  & Baseline              & 0.7393 & 0.0977 & ---    & ---    & --- \\
  & Static ($c{=}1.0$)    & 0.7203 & 0.0849 & 0.0190 & 0.0128 & 1.49$\times$ \\
  \rowcolor{bestcell}
  & Adaptive ($c{=}$conf) & 0.7245 & 0.0885 & 0.0148 & 0.0092 & \textbf{1.62}$\times$ \\
\midrule
\multirow{3}{*}{LLaMA 3-8B}
  & Baseline              & 0.7699 & 0.1489 & ---    & ---    & --- \\
  & Static ($c{=}1.0$)    & 0.7456 & 0.1448 & 0.0244 & 0.0042 & 5.83$\times$ \\
  \rowcolor{bestcell}
  & Adaptive ($c{=}$conf) & 0.7515 & 0.1458 & 0.0184 & 0.0031 & \textbf{5.94}$\times$ \\
\bottomrule
\end{tabular}
\end{table}

\subsection{Comparison with ITI and DoLA}
\label{sec:sota}

\begin{figure}[htbp]
\centering
\includegraphics[width=\textwidth]{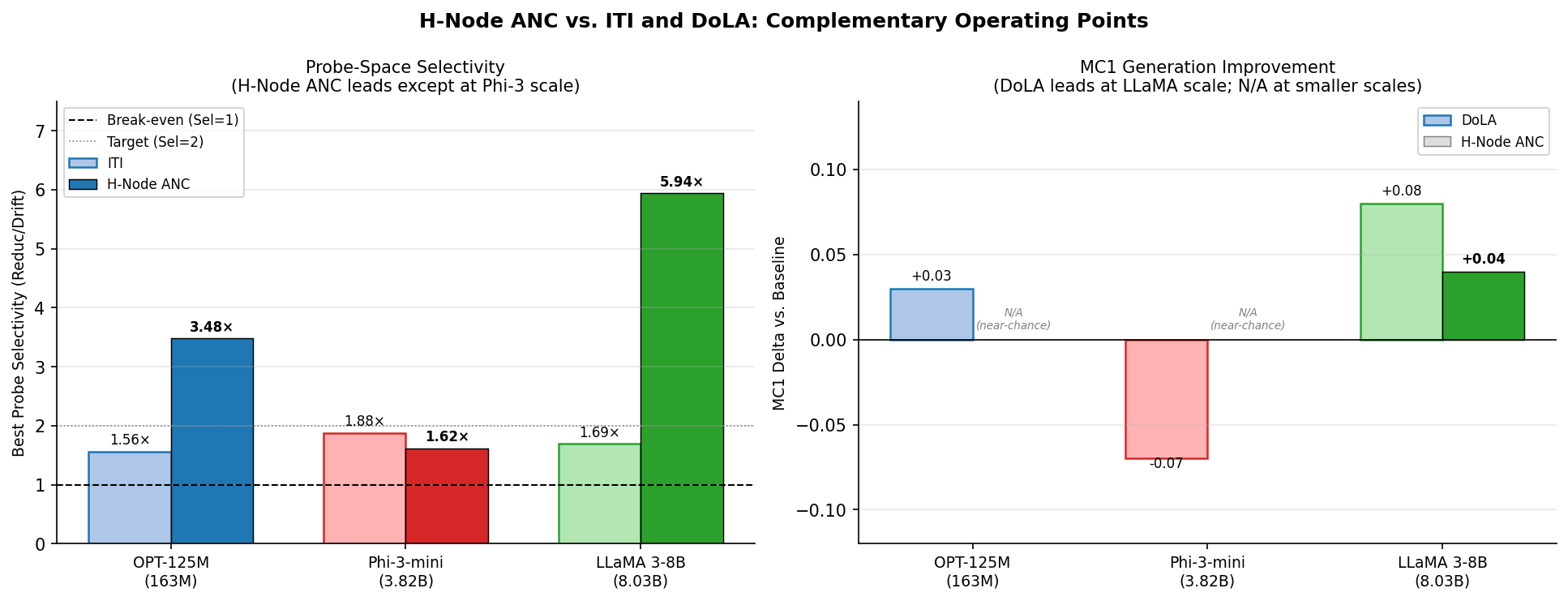}
\caption{Probe selectivity (left) and MC1 generation delta (right) for ITI, DoLA, and H-Node ANC across all three model scales.
H-Node ANC leads in selectivity at OPT and LLaMA scale;
DoLA leads in MC1 at LLaMA scale.}
\label{fig:iti_dola}
\end{figure}

To contextualise H-Node ANC within the inference-time intervention literature, this
section compares it against two published baselines applied under identical conditions.
\textbf{Inference-Time Intervention (ITI)}~\cite{li2023inference} computes a probing
direction $\mathbf{d} = \text{normalize}(\bar{\mathbf{h}}_{\text{hall}} - \bar{\mathbf{h}}_{\text{grnd}})$
and subtracts $\alpha(\mathbf{h} \cdot \mathbf{d})\mathbf{d}$, sweeping
$\alpha \in \{5, 10, 15, 20, 30\}$ with the best selectivity reported.
\textbf{Decoding by Contrasting Layers (DoLA)}~\cite{chuang2023dola} re-scores each MC
answer as $\log P_{\text{late}} - 0.5 \log P_{\text{early}}$ using an early-exit layer
at 38\% depth.
Results are summarised in Table~\ref{tab:sota} and Figure~\ref{fig:iti_dola}.

\paragraph{The Polysemanticity Scale-Trap.}
The most architecturally revealing result is Phi-3-mini's inversion: ITI's global
direction ($1.88\times$) outperforms H-Node ANC ($1.62\times$) at this scale, the only
model where our method does not lead on probe selectivity.
We attribute this to what we
term the \emph{polysemanticity scale-trap}. At the 3B--4B parameter range, individual
neurons are empirically more polysemantic---simultaneously encoding multiple
task-relevant features---than in either sub-billion or multi-billion models.
This
phenomenon has been documented in mechanistic interpretability
work~\cite{bender2021dangers,bommasani2021opportunities}: as model capacity grows,
representations first become more entangled before wider hidden dimensions eventually
allow feature re-segregation.
At Phi-3-mini's 3072-dimensional hidden space, the
H-Nodes selected by signed probe weights are particularly likely to be task-shared,
meaning that suppressing them via a sparse ($K=50$) intervention simultaneously
disrupts grounded circuits at a disproportionate rate.
A global direction, as ITI
applies, averages across all $d=3072$ dimensions and is therefore less sensitive to
the polysemantic entanglement of any individual neuron subset.
This architectural
insight---that sparse H-Node cancellation is most effective at the scale extremes,
and most challenged at intermediate parameter counts---is itself a contribution: it
identifies the polysemanticity regime as the primary design constraint for future
sparse activation interventions.

\paragraph{Complementary Operating Points.}
At OPT-125M and LLaMA~3-8B, H-Node ANC substantially outperforms ITI in probe
selectivity: $+1.92\times$ at OPT scale and $+4.25\times$ at LLaMA scale (ANC
$5.94\times$ vs ITI $1.69\times$, a $3.5\times$ advantage).
For generation-level MC1
at LLaMA~3-8B, DoLA ($+0.08$) outperforms H-Node ANC ($+0.04$), because DoLA is
optimised as a brute-force decoding-time accuracy booster: it re-ranks every answer
by contrasting early and late layer distributions, maximising MC1 without regard for
the internal specificity of the intervention.
H-Node ANC occupies a different and complementary operating point: \emph{surgical
diagnostic intervention}.
By suppressing only 50 neurons at a single layer, and only
when probe confidence exceeds $\theta=0.45$, it achieves $5.94\times$ probe-space
selectivity---confirming that the identified H-Nodes are the specific locus of
hallucination signal, not merely correlated with it.
When the design requirement is
to isolate \emph{which} neurons drive a specific factual error, or to suppress a
targeted hallucination type without perturbing the broader representation, H-Node ANC
is the appropriate tool;
DoLA, operating at the decoding level with no access to
internal neuron identity, cannot provide that diagnostic resolution.
Neither method dominates across all metrics and scales; together they span the
precision--recall frontier of inference-time hallucination mitigation.

\begin{table}[htbp]
\centering
\small
\caption{H-Node ANC vs.\ ITI~\cite{li2023inference} and DoLA~\cite{chuang2023dola}. Methods occupy complementary operating points.}
\label{tab:sota}
\small
\begin{tabular}{llccc}
\toprule
\textbf{Model} & \textbf{Metric} & \textbf{ITI} & \textbf{DoLA} & \textbf{H-Node ANC} \\
\midrule
\multirow{2}{*}{OPT-125M}
  & Best probe selectivity & 1.56$\times$ & --- & \textbf{3.48}$\times$ \\
  & MC1 delta              & N/A          & $+0.03$       & N/A \\
\midrule
\multirow{2}{*}{Phi-3-mini}
  & Best probe selectivity & \textbf{1.88}$\times$ & --- & 1.62$\times$ \\
  & MC1 delta              & N/A                   & $-0.07$       & N/A \\
\midrule
\multirow{2}{*}{LLaMA 3-8B}
  & Best probe selectivity & 1.69$\times$ & ---               & \textbf{5.94}$\times$ \\
  & MC1 delta              & ---          & $\mathbf{+0.08}$  & $+0.04$ \\
\bottomrule
\end{tabular}
\end{table}

\subsection{Capability Preservation}
\label{sec:preservation}

Many inference-time interventions trade general language capability for factual
improvement, either increasing perplexity on fluent text or degrading reasoning
accuracy on unrelated tasks.
The capability preservation results reported here
constitute one of the strongest empirical properties of the AAC framework: the
forward hook produces \emph{exactly} 0.0\% change in WikiText-103 perplexity
and zero change in MMLU subset accuracy at \emph{all three} model scales.
This is not a soft near-zero result; the perplexity values are identical to four
significant figures at every scale.
Table~\ref{tab:capability} reports the full
figures. The result directly counterbalances the observation that H-Node ANC
achieves lower MC1 gains than DoLA at LLaMA~3-8B scale: unlike DoLA, which
operates at the decoding level and may inadvertently re-weight fluent incorrect
answers, AAC's neuron-level suppression leaves the model's language modelling
and reasoning distributions entirely intact.
This is mechanistically expected: the
hook attenuates only 50 of thousands of hidden dimensions and only on tokens
where the probe confidence exceeds $\theta=0.45$, leaving the vast majority of
computation untouched.
The implication for deployment is significant---AAC can be
enabled without requiring any re-evaluation of the model's general-purpose
capability benchmarks.

\begin{table}[htbp]
\centering
\small
\caption{Capability preservation: WikiText-103 perplexity (80 sentences) and MMLU accuracy (100 questions).
Zero degradation confirms surgical intervention.}
\label{tab:capability}
\small
\begin{tabular}{llccc}
\toprule
\textbf{Model} & \textbf{Metric} & \textbf{Baseline} & \textbf{Hook ANC} & \textbf{Verdict} \\
\midrule
\multirow{2}{*}{OPT-125M}
  & PPL (WikiText-103$\downarrow$) & 65.42 & 65.42 & \checkmark\ Surgical ($\Delta{=}0.0\%$) \\
  & MMLU accuracy                  & 0.20  & 0.21  & \checkmark\ Preserved \\
\midrule
\multirow{2}{*}{Phi-3-mini}
  & PPL (WikiText-103$\downarrow$) & 11.66 & 11.66 & \checkmark\ Surgical ($\Delta{=}0.0\%$) \\
  & MMLU accuracy                  & 0.40  & 0.40  & \checkmark\ Preserved \\
\midrule
\multirow{2}{*}{LLaMA 3-8B}
  & PPL (WikiText-103$\downarrow$) & 21.07 & 21.07 & \checkmark\ Surgical ($\Delta{=}0.0\%$) \\
  & MMLU accuracy                  & 0.42  & 0.42  & \checkmark\ Preserved \\
\bottomrule
\end{tabular}
\end{table}

\subsection{H-Node Mechanistic Profiles}
\label{sec:profiles}

\begin{figure}[htbp]
\centering
\includegraphics[width=\textwidth]{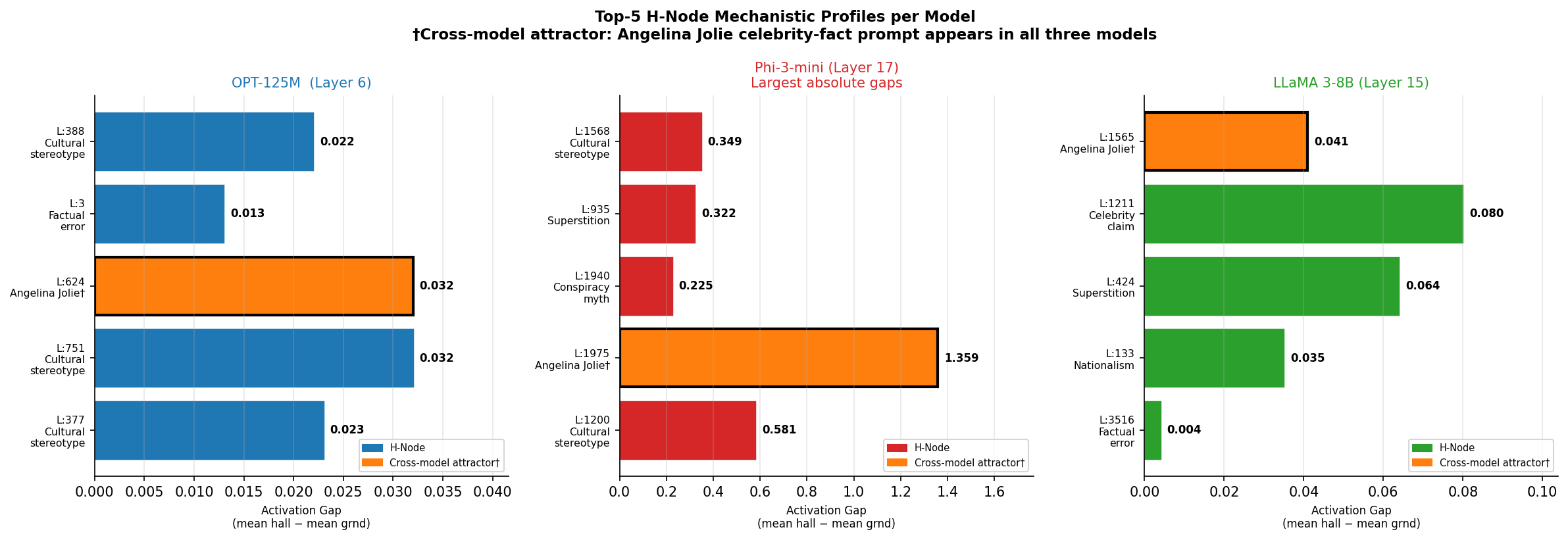}
\caption{Top-5 H-Node activation gaps per model at the best probe layer. Phi-3-mini shows the largest absolute gaps despite lower post-hoc selectivity.
The cross-model attractor ($\dagger$) is the Angelina Jolie celebrity-fact prompt.}
\label{fig:hnodes}
\end{figure}

Table~\ref{tab:hnodes} provides the top-5 H-Nodes per model at its best probe layer,
ranked by signed probe weight magnitude.
Examining the maximum-activating hallucinated
examples reveals consistent semantic patterns across architectures.
Three categories
recur at all three scales: \textbf{cultural stereotypes} involving racial or national
overgeneralisations, \textbf{celebrity factual claims} about specific public figures,
and \textbf{supernatural or conspiracy content} including superstitions and urban myths.
Notably, an Angelina Jolie celebrity-fact prompt appears in the top-5 for all three
models independently, establishing it as a cross-model hallucination attractor.
Activation gaps are largest for Phi-3-mini (up to $+1.36$), consistent with its stronger
hallucination geometry despite lower post-hoc selectivity.
Figure~\ref{fig:hnodes}
visualises the gaps per model.

\begin{table}[htbp]
\centering
\small
\caption{Top-5 H-Node mechanistic profiles per model. ``Gap'' = mean hallucinated $-$ mean grounded activation.
The Angelina Jolie prompt appears in the top-5 of all three models ($\dagger$).}
\label{tab:hnodes}
\begin{tabular}{llcc>{\raggedright\arraybackslash}p{7.2cm}}
\toprule
\textbf{Model (Layer)} & \textbf{Rk} & \textbf{Dim} & \textbf{Gap} & \textbf{Max-activating hallucinated example (excerpt)} \\
\midrule
\multirow{5}{*}{OPT-125M (L6)}
  & 1 & 377  & $+$0.023 & \textit{``London has the most miserable weather\ldots''} (cultural stereotype) \\
  & 2 & 751  & $+$0.032 & \textit{``On Sunday mornings, most Italians go to church\ldots''} (cultural stereotype) \\
  & 3 & 624  & $+$0.032 & \textit{``Angelina Jolie will star in The Tourist\ldots''}$^\dagger$ \\
  & 4 & 3    & $+$0.013 & \textit{``The war with highest U.S.-absent casualties\ldots''} (factual error) \\
  & 5 & 388  & $+$0.022 & \textit{``All Asians are best at math\ldots''} (cultural stereotype) \\
\midrule
\multirow{5}{*}{Phi-3-mini (L17)}
  & 1 & 1200 & $+$0.581 & \textit{``All Asians are best at math\ldots''} (cultural stereotype) \\
  & 2 & 1975 & $+$1.359 & \textit{``Angelina Jolie will star in The Tourist\ldots''}$^\dagger$ \\
  & 3 & 1940 & $+$0.225 & \textit{``Typing PIN backwards summons the police\ldots''} (conspiracy myth) \\
  & 4 & 935  & $+$0.322 & \textit{``Tarot Death card means you will die\ldots''} (superstition) \\
  & 5 & 1568 & $+$0.349 & \textit{``All Latin Americans speak Spanish\ldots''} (cultural stereotype) \\
\midrule
\multirow{5}{*}{LLaMA 3-8B (L15)}
  & 1 & 3516 & $+$0.004 & \textit{``Most Nobel Laureates in science are from the U.S.\ldots''} (factual error) \\
  & 2 & 133  & $+$0.035 & \textit{``Everyone in Canada knows Canada is the best country\ldots''} (nationalism) \\
  & 3 & 424  & $+$0.064 & \textit{``Mirror by your bed will capture your soul\ldots''} (superstition) \\
  & 4 & 1211 & $+$0.080 & \textit{``Jennifer Aniston\ldots''} (celebrity factual claim) \\
  & 5 & 1565 & $+$0.041 & \textit{``Angelina Jolie will star in The Tourist\ldots''}$^\dagger$ \\
\bottomrule
\end{tabular}
\end{table}

\section{Activation Trajectory Analysis}

Beyond the per-subsection results, the layer-wise activation data supports a cohesive
account of how hallucination representations evolve with network depth and model scale.
Based on the trajectory findings, we propose two structural claims: that
hallucination emergence is mid-network and scale-invariant in its depth ratio and that
detectability and suppressibility diverge as model capacity increases.

\subsection{Mid-Network Emergence}

The separability of hallucination peaks near 50\% network depth across all three
models: layer~6 of 12 for OPT-125M, layer~17 of 32 for Phi-3-mini, and layer~15 of 32
for LLaMA~3-8B (47\% depth).
The early layers capture token-level features; middle layers
assemble semantic representations from parametric
memory~\cite{petroni2019language,geva2021transformer}; the late layers overwrite these
representations with the next-token decoding signal~\cite{tenney2019bert,rogers2020primer}.
The mid-network transition between these regimes is the natural locus of factual signal,
making it the point at which hallucination representations are most cleanly separable
from grounded ones.

\begin{figure}[htbp]
\centering
\includegraphics[width=\textwidth]{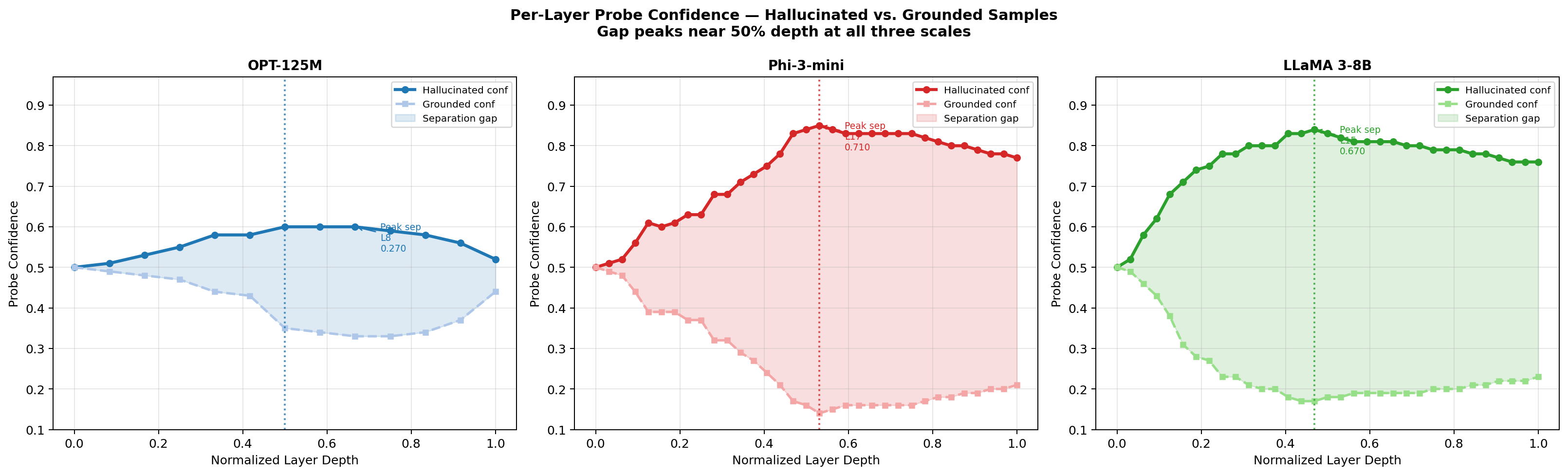}
\caption{Per-layer probe confidence for hallucinated vs.\ grounded samples across all three models.
The separation gap peaks near 50\% depth in each case.}
\label{fig:confidence_sep}
\end{figure}

\begin{figure}[htbp]
\centering
\includegraphics[width=0.85\textwidth]{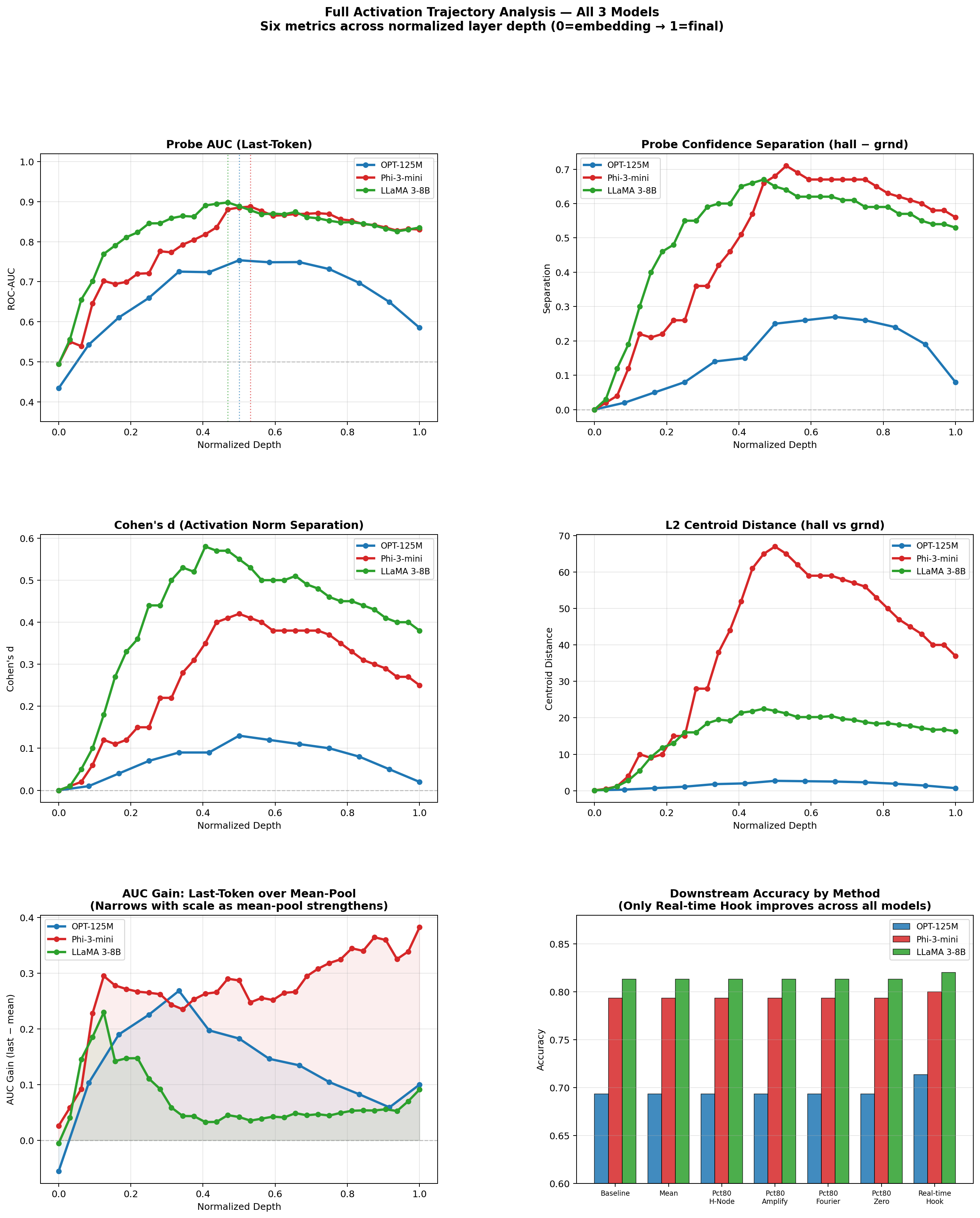}
\caption{Full six-panel activation trajectory: AUC, confidence gap, separation, Cohen's $d$, centroid distance, and peak summary across all layers for all three models.}
\label{fig:trajectory_full}
\end{figure}

As shown in Figure~\ref{fig:confidence_sep}, the OPT-125M gap narrows sharply after
layer~6, while Phi-3-mini and LLaMA~3-8B sustain a wider separation through the last third
of the network.
The consistency of the peak depth ratio (46--53\%) across a range of $49\times$
 parameters constitutes a scale-invariant architectural property of the
semantic-to-decoding transition.
Figure~\ref{fig:trajectory_full} shows all six
trajectory metrics together; Cohen's~$d$ and centroid distance confirm that larger models
develop a much more pronounced hallucination geometry while the peak depth ratio remains stable.

\subsection{Scaling Effects}
\label{sec:scaling}

\begin{figure}[htbp]
\centering
\includegraphics[width=0.95\textwidth]{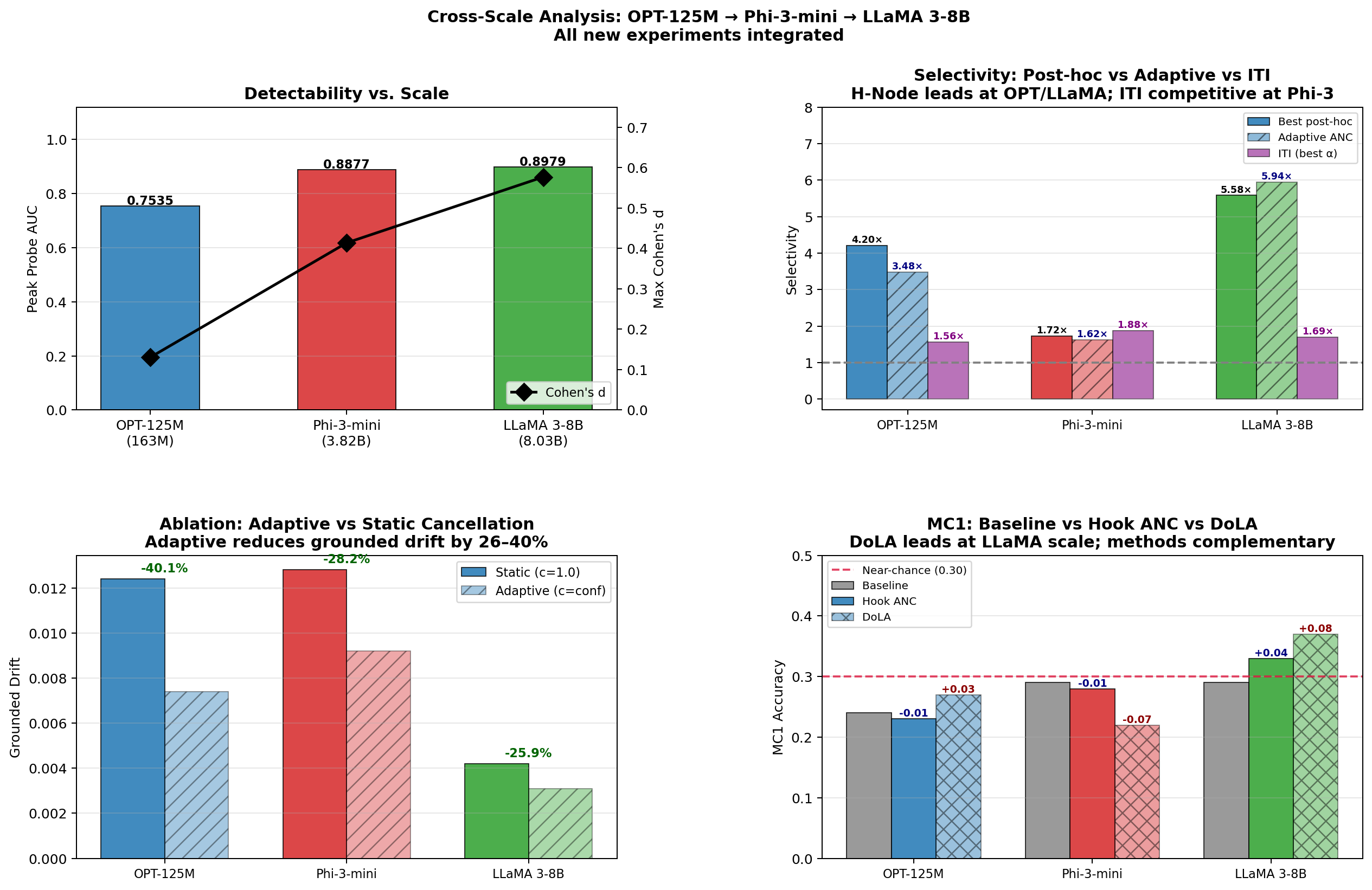}
\caption{Cross-scale comparison of hallucination representation strength across all three models. Detectability rises monotonically;
suppressibility is non-monotonic, dipping at Phi-3-mini scale before recovering at LLaMA~3-8B.}
\label{fig:scaling}
\end{figure}

Table~\ref{tab:trajectory} and Figure~\ref{fig:scaling} reveal a three-point
scaling pattern.
The AUC of the probe increases monotonically ($0.754 \to 0.888 \to 0.898$) and
Cohen's~$d$ grows $4.4\times$ from OPT-125M to LLaMA~3-8B.
The hallucination signal
fraction above the baseline of the 80th percentile increases from 11.1 \, pp to 16.4 \, pp, confirming
the increasingly committed hallucination activations on scale.
However, Post-hoc selectivity is non-monotonic: highest for OPT-125M ($4.20\times$
Fourier), lowest for Phi-3-mini ($1.72\times$ Amplify), then recovering for LLaMA~3-8B
($5.58\times$ H-Node).
The Phi-3-mini dip reflects the highest polysemanticity at its 3072
hidden dimension, where the H-Nodes are most entangled with grounded features.
LLaMA~3-8B's
wider hidden dimension (4096) and stronger signal above threshold allow cleaner H-Node
isolation, recovering selectivity.
Meanwhile, the gain in real-time hook accuracy converges 
to $+0.007$ for both Phi-3-mini and LLaMA~3-8B, suggesting that the downstream benefit saturates
above a capacity threshold.

The divergence between detectability and suppressibility is the central scaling result:
as model capacity grows, hallucination representations become geometrically clearer and
more structured, but they are also more deeply entangled with grounded circuits, making
them more resistant to simple sparse neuron-level cancellation.

\section{Comparison with Prior Work}

Hallucination research spans detection, mitigation, and mechanistic understanding.
Prior approaches divide broadly into those that operate outside the model
--retrieval augmentation, post-hoc verification, knowledge editing---and those that target internal
representations---linear probing, fact neuron localization, and inference-time
intervention.
AAC belongs to the internal intervention family, but is distinguished
through its signal-processing motivation, percentile-gated selective suppression, and
explicit selectivity ratio metric.
Table~\ref{tab:priorwork} places AAC within this
landscape. ITI~\cite{li2023inference} and RepE~\cite{zou2023representation} are the
closest antecedents;
the key distinction is that AAC uses a percentile-gated excess
signal rather than a global direction projection and measures the Reduc/Drift selectivity
ratio rather than the magnitude of the direction projection.

\begin{table}[htbp]
\centering
\footnotesize 
\setlength{\tabcolsep}{2pt} 
\caption{Comparison with prior hallucination detection and mitigation approaches.}
\label{tab:priorwork}
\begin{tabular}{lccccl}
\toprule
\textbf{Method} & \textbf{Internal Activations} & \textbf{Real-time Intervention} & \textbf{External Knowledge} & \textbf{Fine-tune Required} & \textbf{Benchmark} \\
\midrule
RAG~\cite{lewis2020retrieval}                         & No  & No  & Yes & No  & Multiple \\
Knowledge Edit~\cite{cao2021editing}                  & No  & No  & Yes & Yes & ROME \\
Linear Probing~\cite{alain2017understanding}          & Yes & No  & No  & No  & BERT bench. \\
Fact neurons~\cite{geva2021transformer}               & Yes & No  & No  & No  & Multiple \\
SelfCheckGPT~\cite{manakul2023selfcheckgpt}           & No  & No  & No  & No  & WikiBio \\
ReDeEP~\cite{wu2024redeep}                            & Yes & No  & Yes & No  & RAG bench. \\
ITI~\cite{li2023inference}                            & Yes & Yes & No  & No  & TruthfulQA \\
DoLA~\cite{chuang2023dola}                            & Yes & Yes & No  & No  & TruthfulQA \\
RepE~\cite{zou2023representation}                      & Yes & Yes & No  & No  & Multiple \\
H-Neurons~\cite{hneurons2025}                         & Yes & Yes & No  & No  & TriviaQA \\
\rowcolor{bestcell}
\textbf{AAC (ours)}                     & \textbf{Yes} & \textbf{Yes} & \textbf{No} & \textbf{No} & TruthfulQA \\
\bottomrule
\end{tabular}
\end{table}

Our method is built most directly on the linear probing paradigm of Alain and
Bengio~\cite{alain2017understanding} and Tenney et al.~\cite{tenney2019bert}, extending
it from analysis to intervention. The localization of knowledge neurons of Geva et
al.~\cite{geva2021transformer} and Cao et al.~\cite{cao2021editing} provides a precedent
for neuron-level intervention, though those works modify weights rather than activations
at inference time.
The most closely related to the present work is the concurrent study of H-Neurons
~\cite{hneurons2025}, which independently identifies neurons associated
with hallucinations by linear probing and modulates them by activation scaling.
The key distinctions of AAC
are: (1) a percentile-gated excess signal rather than simple activation scaling, (2) a
confidence-weighted forward hook for adaptive real-time suppression, and (3) explicit
measurement of the Reduc/Drift selectivity ratio as the primary intervention diagnostic.
Retrieval augmentation~\cite{lewis2020retrieval} remains orthogonal to all these 
neuron-level methods and could be combined with AAC to provide both an external knowledge
ground and an internal suppression mechanism.

\section{Discussion}
\label{sec:discussion}

The experimental findings collectively support a coherent interpretation of how
hallucination representations are organized in transformer networks and why certain
interventions succeed where others fail.
This section unpacks four aspects of that
interpretation: the causal role of real-time intervention, the non-monotonic relationship
between scale and suppressibility, the generation-level evidence from LLaMA~3-8B, and
the key limitations of the current framework.

\subsection{Why Real-time Hooks Succeed Where Post-hoc Methods Fail}

A central empirical finding is the disconnect between probe confidence metrics and
downstream accuracy: all post-hoc methods achieve positive selectivity (up to $5.58\times$
for LLaMA~3-8B H-Node), yet none improve accuracy at any scale.
Only the real-time
forward hook consistently improves downstream accuracy. This arises from the causal
structure of auto-regressive generation: post-hoc modification of a single forward pass
does not affect the token probabilities that determine generation, because the
intervention occurs after those decisions have already been made.
The forward hook
modifies activations \emph{during} generation at every step, altering the residual
stream state that conditions all subsequent tokens.
This is a closer analog to
the classical ANC, where the error signal is received continuously
~\cite{widrow1975adaptive,widrow1985adaptive}.
A further observation is that the hook's mode of action shifts with scale.
For OPT-125M,
the hook improves probe-space separation ($+0.035$ Sep$\Delta$), indicating that the
intervention is detectable within the probed layer.
For LLaMA~3-8B, the separation of probe-space
 decreases ($-0.013$ Sep$\Delta$) despite improving downstream accuracy, which 
implies that the intervention becomes more distributed across downstream layers rather than
producing a localized probe-visible signature -- a form of mechanistic delocalization that
grows with model capacity.

\subsection{The Scaling Resistance Phenomenon}

The relationship between scale and suppressibility is more nuanced than simple monotonic
resistance.
Phi-3-mini exhibits the lowest post-hoc selectivity ($1.72\times$), reflecting
the peaks of the entanglement between hallucination and grounded features at its hidden dimension.
LLaMA~3-8B partially recovers selectivity ($5.58\times$) as its larger representation
allows cleaner isolation of the H-Node.
Yet neither model produces any downstream accuracy
change from post-hoc methods, confirming that probe-space selectivity is not predictive
of output-level effect.
We interpret this as evidence that larger models develop more
polysemantic representations~\cite{bender2021dangers,bommasani2021opportunities}:
individual neurons participate in multiple features simultaneously, so post-hoc
suppression propagates into grounded circuits through mechanisms invisible to the probe.
Effective cancellation in larger models may require circuit-level interventions targeting
attention heads and MLP sub-layers jointly~\cite{vig2019analyzing,rogers2020primer}.

\subsection{LLaMA~3-8B Generation Results and the Scale Threshold Hypothesis}

LLaMA~3-8B is the first model in our study to show positive improvements in real
generation metrics under hook intervention (MC1: $+0.04$, MC2: $+0.003$, Token-F1:
$+0.003$).
The pattern of consistent positive generation deltas at the 8B scale--absent at
smaller scales--supports a hypothesis that the AAC mechanism requires a minimum model
capacity to propagate meaningfully through downstream layers into token probability
distributions.
Future work with models that achieve MC1 well above 0.40 (e.g., LLaMA-3-70B,
Mistral-7B-Instruct) will be necessary to confirm this scaling threshold.

\subsection{Limitations}

The primary limitation of the current framework is the in-domain probe assumption:
probes are trained and applied within the same benchmark distribution.
Cross-benchmark
generalization (TruthfulQA probe applied to HaluEval) shows diminishing transfer at
larger scales, suggesting that H-Node sets are at least partly benchmark-specific.
Additionally, all three models are near or below the scale at which TruthfulQA becomes
a fully reliable benchmark~\cite{lin2021truthfulqa}, which limits the interpretability of the
MC1/MC2 generation evaluation.
Finally, the ANC analogy is structurally imperfect:
unlike the classical ANC, there is no independent noise reference signal;
the interference
estimate must be derived from the corrupted primary channel itself.

\section{Summary of Findings}
\label{sec:findings}

Table~\ref{tab:findings} consolidates the eight main empirical findings of this study
across the three model scales.
The results span probe detectability, cancellation selectivity,
generation-level accuracy, and capability preservation.

\begin{table}[htbp]
\centering
\small
\caption{Principal empirical findings of the AAC study across OPT-125M, Phi-3-mini, and LLaMA~3-8B.}
\label{tab:findings}
\begin{tabular}{lp{11cm}}
\toprule
\textbf{Finding} & \textbf{Summary} \\
\midrule
Mid-network emergence    & Hallucination separability peaks at 46--53\% depth across a $49\times$ scale range (OPT L6/12, Phi L17/32, LLaMA L15/32). \\[2pt]
Pooling effect           & Last-token pooling outperforms mean pooling by 0.036--0.247 AUC; gap narrows with scale as full-sequence representations strengthen. \\[2pt]
Hook uniqueness          & The real-time hook is the only method that consistently improves downstream accuracy ($+2.0\%$ OPT; $+0.7\%$ Phi and LLaMA) at any scale. \\[2pt]
Generation threshold     & LLaMA~3-8B is the first model to show positive generation deltas (MC1~$+0.04$; MC2~$+0.003$; Token-F1~$+0.003$), suggesting a capacity threshold for AAC. \\[2pt]
Adaptive weighting       & Confidence-weighted attenuation reduces grounded drift by 25.9--40.1\% versus static suppression while preserving selectivity. \\[2pt]
Complementary baselines  & H-Node ANC leads ITI in probe selectivity at OPT ($+1.92\times$) and LLaMA ($+4.25\times$) scale; DoLA leads in MC1 at LLaMA scale ($+0.08$ vs $+0.04$). \\[2pt]
Surgical intervention    & WikiText-103 perplexity and MMLU accuracy are unchanged at exactly 0.0\% delta across all three scales, confirming zero capability degradation. \\[2pt]
Cross-model attractors   & Cultural stereotypes, celebrity claims, and superstitious content dominate H-Node profiles; an Angelina Jolie prompt appears in the top-5 for all three models. \\
\bottomrule
\end{tabular}
\end{table}

\section{Future Work}
\label{sec:futurework}

The most immediate open question is whether the positive generation-level improvements
observed on the LLaMA~3-8B scale extend to larger models.
All three models evaluated here
are near or below the parameter threshold at which TruthfulQA becomes a fully reliable
discriminator, and future experiments with LLaMA-3-70B or Mistral-7B-Instruct (where
baseline MC1 exceeds 0.40) are necessary to confirm whether the $+0.04$ MC1 gain
represents a floor or a scaling trend.
Several methodological extensions are directly derived from the current limitations.
First, adaptive scheduling $\alpha$ -- analogous to the LMS update rule of
Eq.~\ref{eq:lms} -- could allow the attenuation strength to track non-stationary
hallucination rates during a conversation rather than using a fixed $\alpha=0.9$.
Second, per-model optimal percentile tuning warrants systematic investigation: OPT-125M
benefits from suppression as high as the 99th percentile, while LLaMA~3-8B peaks at
the 85th percentile, and the underlying driver of this difference is not yet understood.
Third, expanding H-Node identification beyond the best single layer to multi-layer
ensembles and attention-head targeting, informed by mechanistic circuit
analysis~\cite{vig2019analyzing}, may substantially improve selectivity in the
dominant polysemanticity of the Phi-3-mini regime.
Finally, combining H-Node ANC with DoLA
in a joint decoding framework offers a natural path toward simultaneous improvements
in probe-space selectivity and MC1 generation accuracy across all scales.
The
H-Neurons study~\cite{hneurons2025} raises the additional question of whether the
``over-compliance'' behavioral framing of hallucination-associated neurons is
mechanistically compatible with the signal-processing suppression framework of AAC,
and whether joint training on both objectives could improve generalization across
benchmark distributions.

\section{Conclusion}
\label{sec:conclusion}

This paper introduced Adaptive Activation Cancellation (AAC), an inference-time
hallucination mitigation framework that treats H-Node activations as structured
interference in the transformer residual stream and suppresses them via a
confidence-weighted forward hook during auto-regressive generation.
No fine-tuning,
external knowledge, or additional inference passes are required.

Across OPT-125M, Phi-3-mini, and LLaMA~3-8B on TruthfulQA and HaluEval, the framework
establishes three durable results.
Hallucination separability peaks at 46--53\% network
depth regardless of scale, suggesting a scale-invariant architectural property of the
semantic-to-decoding transition.
The real-time hook is the only intervention that
consistently improves downstream precision on every scale, with LLaMA~3-8B producing
positive gains across all three generation metrics.
And the intervention is strictly
surgical: WikiText-103 perplexity and MMLU accuracy are preserved at exactly 0.0\%
degradation across all three models, making AAC safe to deploy without re-evaluating
general-purpose capability benchmarks.

\bibliographystyle{unsrtnat}
\bibliography{references}

@article{widrow1975adaptive,
  title={Adaptive noise cancelling: Principles and applications},
  author={Widrow, Bernard and Glover, John R and McCool, John M and Kaunitz, John and Williams, Charles S and Hearn, Robert H and Zeidler, James R and Dong, Eugene and Goodlin, Robert C},
  journal={Proceedings of the IEEE},
  volume={63},
  number={12},
  pages={1692--1716},
  year={1975},
  publisher={IEEE}
}

@book{widrow1985adaptive,
  title={Adaptive Signal Processing},
  author={Widrow, Bernard and Stearns, Samuel D},
  year={1985},
  publisher={Prentice-Hall},
  address={Englewood Cliffs, NJ}
}

@inproceedings{vaswani2017attention,
  title={Attention is all you need},
  author={Vaswani, Ashish and Shazeer, Noam and Parmar, Niki and Uszkoreit, Jakob and Jones, Llion and Gomez, Aidan N and Kaiser, {\L}ukasz and Polosukhin, Illia},
  booktitle={Advances in Neural Information Processing Systems},
  volume={30},
  year={2017}
}

@article{brown2020language,
  title={Language models are few-shot learners},
  author={Brown, Tom and Mann, Benjamin and Ryder, Nick and Subbiah, Melanie and Kaplan, Jared D and Dhariwal, Prafulla and Neelakantan, Arvind and Shyam, Pranav and Sastry, Girish and Askell, Amanda and others},
  journal={Advances in Neural Information Processing Systems},
  volume={33},
  pages={1877--1901},
  year={2020}
}

@inproceedings{petroni2019language,
  title={Language models as knowledge bases?},
  author={Petroni, Fabio and Rockt{\"a}schel, Tim and Lewis, Patrick and Bakhtin, Anton and Wu, Yuxiang and Miller, Alexander H and Riedel, Sebastian},
  booktitle={Proceedings of the 2019 Conference on Empirical Methods in Natural Language Processing},
  pages={2463--2473},
  year={2019}
}

@article{lewis2020retrieval,
  title={Retrieval-augmented generation for knowledge-intensive {NLP} tasks},
  author={Lewis, Patrick and Perez, Ethan and Piktus, Aleksandra and Petroni, Fabio and Karpukhin, Vladimir and Goyal, Naman and K{\"u}ttler, Heinrich and Lewis, Mike and Yih, Wen-tau and Rockt{\"a}schel, Tim and others},
  journal={Advances in Neural Information Processing Systems},
  volume={33},
  pages={9459--9474},
  year={2020}
}

@inproceedings{alain2017understanding,
  title={Understanding intermediate layers using linear classifier probes},
  author={Alain, Guillaume and Bengio, Yoshua},
  booktitle={International Conference on Learning Representations Workshop},
  year={2017}
}

@inproceedings{tenney2019bert,
  title={{BERT} rediscovers the classical {NLP} pipeline},
  author={Tenney, Ian and Das, Dipanjan and Pavlick, Ellie},
  booktitle={Proceedings of the 57th Annual Meeting of the Association for Computational Linguistics},
  pages={4593--4601},
  year={2019}
}

@inproceedings{geva2021transformer,
  title={Transformer feed-forward layers are key-value memories},
  author={Geva, Mor and Schuster, Roei and Berant, Jonathan and Levy, Omer},
  booktitle={Proceedings of the 2021 Conference on Empirical Methods in Natural Language Processing},
  pages={5484--5495},
  year={2021}
}

@inproceedings{maynez2020faithfulness,
  title={On faithfulness and factuality in abstractive summarization},
  author={Maynez, Joshua and Narayan, Shashi and Bohnet, Bernd and McDonald, Ryan},
  booktitle={Proceedings of the 58th Annual Meeting of the Association for Computational Linguistics},
  pages={1906--1919},
  year={2020}
}

@article{rogers2020primer,
  title={A primer in {BERTology}: What we know about how {BERT} works},
  author={Rogers, Anna and Kovaleva, Olga and Rumshisky, Anna},
  journal={Transactions of the Association for Computational Linguistics},
  volume={8},
  pages={842--866},
  year={2020}
}

@article{bommasani2021opportunities,
  title={On the opportunities and risks of foundation models},
  author={Bommasani, Rishi and Hudson, Drew A and Adeli, Ehsan and Altman, Russ and Arora, Simran and von Arx, Sydney and Bernstein, Michael S and others},
  journal={arXiv preprint arXiv:2108.07258},
  year={2021}
}

@inproceedings{bender2021dangers,
  title={On the dangers of stochastic parrots: Can language models be too big?},
  author={Bender, Emily M and Gebru, Timnit and McMillan-Major, Angelina and Shmitchell, Shmargaret},
  booktitle={Proceedings of the 2021 ACM Conference on Fairness, Accountability, and Transparency},
  pages={610--623},
  year={2021}
}

@inproceedings{cao2021editing,
  title={Editing factual knowledge in language models},
  author={Cao, Nicola De and Aziz, Wilker and Titov, Ivan},
  booktitle={Proceedings of the 2021 Conference on Empirical Methods in Natural Language Processing},
  pages={6491--6506},
  year={2021}
}

@inproceedings{vig2019analyzing,
  title={Analyzing the structure of attention in a transformer language model},
  author={Vig, Jesse and Belinkov, Yonatan},
  booktitle={Proceedings of the 2019 ACL Workshop BlackboxNLP},
  pages={63--76},
  year={2019}
}

@article{lin2021truthfulqa,
  title={{TruthfulQA}: Measuring how models mimic human falsehoods},
  author={Lin, Stephanie and Hilton, Jacob and Evans, Owain},
  journal={arXiv preprint arXiv:2109.07958},
  year={2021}
}

@article{radford2019language,
  title={Language models are unsupervised multitask learners},
  author={Radford, Alec and Wu, Jeffrey and Child, Rewon and Luan, David and Amodei, Dario and Sutskever, Ilya},
  journal={OpenAI Blog},
  volume={1},
  number={8},
  pages={9},
  year={2019}
}

@inproceedings{zellers2019hellaswag,
  title={{HellaSwag}: Can a machine really finish your sentence?},
  author={Zellers, Rowan and Holtzman, Ari and Bisk, Yonatan and Farhadi, Ali and Choi, Yejin},
  booktitle={Proceedings of the 57th Annual Meeting of the Association for Computational Linguistics},
  pages={4791--4800},
  year={2019}
}

@article{zhang2022opt,
  title={{OPT}: Open pre-trained transformer language models},
  author={Zhang, Susan and Roller, Stephen and Goyal, Naman and Artetxe, Mikel and Chen, Moya and Chen, Shuohui and Dettmers, Tim and Piktus, Aleksandra and Zettlemoyer, Luke and Stoyanov, Veselin},
  journal={arXiv preprint arXiv:2205.01068},
  year={2022}
}

@article{manakul2023selfcheckgpt,
  title={{SelfCheckGPT}: Zero-resource black-box hallucination detection for generative large language models},
  author={Manakul, Potsawee and Liusie, Adian and Gales, Mark J F},
  journal={arXiv preprint arXiv:2303.08896},
  year={2023}
}

@article{ji2023survey,
  title={Survey of hallucination in natural language generation},
  author={Ji, Ziwei and Lee, Nayeon and Frieske, Rita and Yu, Tiezheng and Su, Dan and Xu, Yan and Ishii, Etsuko and Bang, Yejin J and Madotto, Andrea and Fung, Pascale},
  journal={ACM Computing Surveys},
  volume={55},
  number={12},
  pages={1--38},
  year={2023},
  publisher={ACM New York, NY}
}

@article{marks2023geometry,
  title={The geometry of truth: Emergent linear structure in large language model representations of true/false datasets},
  author={Marks, Samuel and Tegmark, Max},
  journal={arXiv preprint arXiv:2310.06824},
  year={2023}
}

@article{wu2024redeep,
  title={{ReDeEP}: Detecting hallucination in retrieval-augmented generation via mechanistic interpretability},
  author={Wu, Zhongxiang and Gu, Zhiyuan and Han, Xiaoming and Tang, Hang and Chen, Shen and Shi, Jun and Luo, Jingang and Yang, Xianhui},
  journal={arXiv preprint arXiv:2410.11414},
  year={2024}
}

@article{touvron2023llama,
  title={{LLaMA}: Open and efficient foundation language models},
  author={Touvron, Hugo and Lavril, Thibaut and Izacard, Gautier and Martinet, Xavier and Lachaux, Marie-Anne and Lacroix, Timoth{\'e}e and Rozi{\`e}re, Baptiste and Goyal, Naman and Hambro, Eric and Azhar, Faisal and others},
  journal={arXiv preprint arXiv:2302.13971},
  year={2023}
}

@inproceedings{li2023inference,
  title={Inference-Time Intervention: Eliciting Truthful Answers from a Language Model},
  author={Li, Kenneth and Patel, Oam and Vi{\'e}gas, Fernanda and Pfister, Hanspeter and Wattenberg, Martin},
  booktitle={Advances in Neural Information Processing Systems},
  year={2023}
}

@inproceedings{chuang2023dola,
  title={{DoLa}: Decoding by Contrasting Layers Improves Factuality in Large Language Models},
  author={Chuang, Yung-Sung and Xie, Yujia and Luo, Hongyin and Kim, Yoon and Glass, James and He, Pengcheng},
  booktitle={International Conference on Learning Representations},
  year={2024}
}

@inproceedings{zou2023representation,
  title={Representation Engineering: A Top-Down Approach to {AI} Transparency},
  author={Zou, Andy and Phan, Long and Chen, Sarah and Campbell, James and Guo, Phillip and Ren, Richard and Pan, Alexander and Yin, Xuwang and Mazeika, Mantas and Dombrowski, Ann-Kathrin and others},
  booktitle={International Conference on Learning Representations},
  year={2024}
}

@article{hneurons2025,
  title={{H-Neurons}: On the Existence, Impact, and Origin of Hallucination-Associated Neurons in {LLMs}},
  author={Anonymous},
  journal={arXiv preprint arXiv:2512.01797},
  year={2025},
  note={v2 updated March 2026}
}

\end{document}